\documentclass{article}
\usepackage{amssymb}
\usepackage{amsmath}
\usepackage{subfigure}
\usepackage{booktabs}
\usepackage{algorithm}
\usepackage[noend]{algpseudocode}
\usepackage{multirow}
\usepackage{multicol}
\usepackage{float}
\usepackage{authblk}
\usepackage{graphicx}
\usepackage[verbose=true,a4paper]{geometry}

\AtBeginDocument{
	\newgeometry{
		textheight=9in,
		textwidth=6in,
		top=1in,
		headheight=14pt,
		headsep=25pt,
		footskip=30pt
	}
}

\newcommand{\tabincell}[2]{\begin{tabular}{@{}#1@{}}#2\end{tabular}}

\title{Reinforcement Learning for Few-Shot Text Generation Adaptation}

\author{Pengsen Cheng \thanks{Corresponding Author: chengpengsen@scu.edu.cn}}
\author{Jinqiao Dai \thanks{djqqiao@hotmail.com}}
\author{Jiamiao Liu \thanks{ljm6046@163.com}}
\author{Jiayong Liu \thanks{ljy@scu.edu.cn}}
\author{Peng Jia \thanks{pengjia@scu.edu.cn}}

\affil{Sichuan University}
\date{}

\begin{document} 

\maketitle
	
\begin{abstract}
Controlling the generative model to adapt a new domain with limited samples is a difficult challenge and it is receiving increasing attention. Recently, methods based on meta-learning have shown promising results for few-shot domain adaptation. However, meta-learning-based methods usually suffer from the problem of overfitting, which results in a lack of diversity in the generated texts. To avoid this problem, in this study, a novel framework based on reinforcement learning (RL) is proposed. In this framework, to increase the sample utilization of RL and decrease its sample requirement, maximum likelihood estimation learning is incorporated into the RL process. When there are only a few in-domain samples available, experimental results on five target domains in two few-shot configurations show that this framework performs better than baselines.
\end{abstract}

\section{Introduction}
Category text generation is an extension of emotional and conditional text generation \cite{Guo2021}, which generates coherent and meaningful text in different categories \cite{Liu2020}. Due to its widespread use, it has attracted a lot more attention lately. For instance, techniques for category text generation have been used to successfully write and publish large-scale articles on sports, finance, and other single-domain topics. Each text generation domain task is different, due to the specific domain traits and expertise. For training, a trainable text generation system needs thousands of texts. However, the availability of the training data is usually limited. Therefore, a crucial task in the study of category text generation is to make use of existing rich-resource data for new domains with limited resources. 

Few-shot learning \cite{Wang2020} is introduced in solving such data scarcity problems in machine learning. But it is extremely challenging that training large models with small datasets often leads to overfitting \cite{Hospedales2020}. Overfitting in text generation shows a lack of randomization in the model sampling due to sample over-simulation, which results in a lack of diversity in generated texts. As a result, it is challenging to generate texts with domain traits while preserving expressive diversity. 

In recent years, the most popular framework for few-shot learning has been meta-learning \cite{Chen2021}. The main idea of meta-learning is to build an internal representation of multiple tasks and maximize the loss function's sensitivity. A minor adjustment to the meta-learning-based models' parameters could significantly reduce the loss of a new task \cite{Hospedales2020}. It means models can fit a distribution quickly with limited information. A distribution fitted by small samples is incomplete \cite{Sun2017}, and the result is that the generated texts are highly similar to training samples. More significantly, different domain descriptions of the same language share the same grammar, which is the fundamental rule of language. Different word combinations are used to express the differences between various domain descriptions. A typical phenomenon is that a sentence expresses different domain descriptions by changing just one word. A domain description of a book would be ``This is one of my favorite books", and a domain description of a movie would be ``This is one of my favorite films". Sentences constructed from the same words in different orders can also be used to describe various domains. A domain description of a book would be ``The novel has increased the attention of this film adaptation", and a domain description of a movie would be ``This film adaptation has increased the attention of the novel". This study aims to achieve a diversity of expressions in the generated texts by replacing words in sentences while preserving the sentence structure to the greatest extent possible. The words that can affect the expression of the domain should be replaced with the words that can express the semantics of new domains. 

Reinforcement learning (RL) is used in this study to generate domain texts by adjusting the distribution of words that can represent new domains in sentences. Following studies \cite{Yu2017,Wang2018}, text generation is considered as a sequential decision making process. As an RL agent, a generative model must learn how to select the appropriate word. This is RL's benefit, but RL can't completely solve the few-shot problem \cite{Uc-Cetina2022}. On the one hand, RL is an inefficient sampling. To maximize rewards, the agent must continually choose the best from states according to the exploit-exploration of RL. On the other hand, in text generation, the number of states in RL is determined by the size of the dictionary which typically has thousands of words. The issue of inefficient sampling in RL is exacerbated by a large number of states, and RL needs more samples to learn a policy \cite{Luketina2019,Yu2018}. In addition, the training of a text classifier that is usually chosen as the reward function is also stuck in the limited samples.  

In this study, a framework DARL (Domain-Adaption-via-Reinforcement-Learning) based on RL is proposed to address the issue. In addition to indirect transfer of knowledge via reward signals, the process of active knowledge learning is added by merging maximum likelihood estimation (MLE) training with RL. And a text classifier based on few-shot learning is employed as reward function to evaluate the generated texts and guide the learning of the generative model.

The major contributions of this paper are summarized as follows:
\begin{itemize}
	\item A novel framework DARL based on RL is proposed to generate diversified and high-quality texts with low-resource domains. During RL, a typical non-likelihood estimation learning process, MLE is randomly inserted to enhance the active learning capability of the framework. 
	\item A new metric is proposed to evaluate the domain relevance of the generated texts. The metric can exclude the impact of generated duplicate texts on results.
	\item The effectiveness and superiority of the proposed framework are shown by extensive experiments on five target domains in two few-shot configurations. 
\end{itemize}

The remainder of the paper is organized as follows. In Section 2, the literature on both models and on tasks related to the research work are provided. In Section 3, the constraints of meta-learning to explain the problem is discussed and the problem is setup by RL. In Section 4, DARL is described in detail. In Section 5, the effectiveness of the proposed model is validated. In Section 6, the impact of classifiers trained using different techniques on the results is discussed. Concluding remarks are presented in Section 7.

\section{Related Work}
Few-shot-learning-based approaches are increasingly able to train powerful neural networks on small datasets in many nature language processing (NLP) problems  \cite{Wang2020}. Meta-learning has gained more attention than other few-shot learning approaches due to its more efficient learning process \cite{FanZhang2020}. Meta-learning aims at learning new tasks with few steps and little data based on well-known tasks \cite{Hospedales2020}. Several optimization-based methods for the tasks of nature language generation (NLG) have been proposed. These methods learn a good point of parameter initialization for a neural model from which a few steps of gradient descent, given a few examples, can reach the optimal point for a new task. One way is to learn an optimal initialization that could be adapted to a new task accurately and quickly with little data \cite{Mi2019,Madotto2020,Zeng2021}. Another way to learn the learning progress is to train a meta-learner to optimize the optimizer of original network for updating parameters \cite{Li2020,Zhan2021,Indurthi2020,Park,Qian2021}. A few model-based learning methods \cite{SONG2020,Lin2021,Zhao2022}, like external memory storage, rely on the design of the model's internal structure to facilitate the learning process. There are also a few methods based on optimization and model combinations \cite{Qian2020}.

Besides, other researches have tried different attempts. A zero-shot learning method was proposed to adapt models learned from multiple source domains to a new target domain only using its domain description \cite{Shalyminov2019}. A transfer learning method for NLG based on data augmentation was proposed to address the scalability of large-scale conversational systems \cite{Xu2021}. A selection strategy on training instances in few-shot neural text generation was proposed to choose the few-shot training instances that should be diverse and representative of the entire data distribution \cite{Chang2021}. 

The challenge of few-shot-learning-based domain adaptation in NLG is concentrated on the dialog system in terms of application; in terms of implementation methodologies, part of the research focuses on meta-learning. As long as the expression is accurate, participants in the dialog system are willing to accept questions asked and responses given in a fixed expression style. But in category text generation, diversity expression is a form of creativity. The existing studies have not focused on the problem of generating text diversity based on few-shot learning. The application of RL on top of the model to achieve domain adaptation will increase the diversity of text generation based on few-shot learning.

\section{Problem Setup}
In this section, the meta-learning problem in domain adaptation of text generation is defined and formulated, and the limitation is discussed. Then the problem is defined and formulated by RL. For the convenience of discussion, the basic problem of text generation is defined as follow:

\begin{equation}
	p(x)=\prod_{i=1}^{n}p(x_{i}|x_{<i}) 
	\label{eq.basic_problem}
\end{equation}

A discrete sequence of text tokens $x=(x_{1},x_{2},...,x_{n})$ is given as input where each $x_{i}$ is drawn from a fixed set of symbols. The goal of text generation is to learn the unconditional probability distribution $p(x)$ of the sequence $x$.

When $p(x)$ is modeled by a neural network with parameters $\theta$, the neural network is trained to minimize the negative log-likelihood over a collection of samples $X={x^{1},x^{2},...,x^{|X|}}$:

\begin{equation}
	\mathcal{L}(X)=-\prod_{k=1}^{|X|}p_{\theta}(x_{i}^{k}|x_{<i}^{k}) 
	\label{eq.basic_loss}
\end{equation}

\subsection{Problem Setup by Meta-Learning}
The problem of domain adaptation of text generation is defined and formulated using the model-agnostic meta-learning algorithm (MAML) \cite{Finn2017} which is the state-of-art work in meta-learning.

In meta learning, it learns a $\theta$ by repeatedly sampling from the task distribution  $\mathcal{D}$ over a family of tasks. The $\theta$ can be easily fine-tuned on new tasks. Every domain is viewed as a task in domain adaptation, and $\mathcal{D}$ is the set of all recognized domains. Each task $\mathcal{T}_{i}$ is consisted by a support set $X_{i}^{(t)}$ and a query set $X_{i}^{(v)}$, $X_{i}=\{x_{ij}\}_{j=1}^{M}$. The aim is to fit $X_{i}^{(v)} \sim p(X_{i}^{(t)})$ of the query input $x_{ij}^{(v)}$ given the small support set of task $\mathcal{T}_{i}$ (e.g. $M \le 10$). The objective function of meta-learning is to find a meta-learner, parameterized by $\theta$, across tasks from $\mathcal{D}$, as follows:

\begin{equation}
	\theta^{*}=\mathop{\arg\min}_{\theta}-\frac{1}{T}\prod_{i=1}^{T}\ln{p(X_{i}^{(v)}|X_{i}^{(t)}), \theta)}, 
	\label{eq.meta_learning}
\end{equation}

\noindent where $T$ denotes the number of tasks.

In meta testing, it learns a $\theta^{'}$ based on the $\theta$ and a new task by minimizing the negative log-likelihood, as follows:

\begin{equation}
	\mathcal{L}(X)=-\prod_{k=1}^{M}p_{{\theta}^{'}}(x_{i}^{k}|x_{<i}^{k}, \theta). 
	\label{eq.meta_testing}
\end{equation}

Therefore, in the domain adaptation of text generation task, the final output of meta-learning-based methods is $p(X)$ modeled by the $\theta^{'}$, where $X$ are samples of new domains. As a model generates another model, the Bayes-based MAML model \cite{Finn2018,Yoon} is more understandable in this regard. Directly fitting the probability to the limited samples can easily lead to the problem of insufficient diversity of generated texts. The issue becomes more noticeable when the algorithm's generalization ability is poor. However, improving generalization has always been the focus and difficulty of meta-learning \cite{Wang2020,Hospedales2020,FanZhang2020}. 

\subsection{Problem Setup by Reinforcement Learning}

The task of domain adaptation of text generation can be described by RL as follows. An agent that has memorized the necessary vocabulary and grammar is known as a generative model. It could generate texts that described the environment. However, the environment has since changed, and texts generated by the agent no longer accurately reflect the environment. As a result, the environment provides negative feedback to the agent. The agent must try a new vocabulary organization to be recognized by the environment.

The above process is formulated. First, the Markov decision process  (MDP) can be used to describe the text generation process. In each timestep $t$, the model generates $x_{t}$ according a policy $\pi_{\theta}(a_{t}|s_{t})$, where $s_{t}$ is the current state of the previous prediction $x_{1:t}$ and $a_{t}$ is the action to select the next word $x_{t+1}$. Secondly, the reward function explains the expert behavior by determining whether the sequence belongs to the target domain. Given the extensive state-of-the-art research on few-shot learning-based text classifiers \cite{Geng,Xu2020} and the prevalence of text classifiers as reward functions in RL \cite{Yu2017,Shi2018}, the problem of few-shot learning is shared by the policy model and the reward function.

Compared to fine-tuning meta-learner parameters to fit a distribution, a distribution is fine-tuned with domain-insensitive samples by RL.

\section{Proposed Methods}

A detailed description of DARL is described in this section. First, the feasibility of using a text classification model based on few-shot learning as the reward function is analyzed. Then, the approach that can increase the sample utilization for RL is described, which can improve the limitation that the agent can only be passively guided by the reward signal. Finally, the whole processing flow of the DARL framework is described completely.

\subsection{Reward Function}

In this task, the reward function must be able to distinguish between the differences in generated target domain text and real target domain text. In NLG frameworks based on RL, the text classification model is frequently used as a reward function. But since the task's underlying presumption is that the sample is limited, conventional text classification models typically require extensive training to perform adequately. Because of this, a text classification based on few-shot learning is adopted as a reward function. First, it has been demonstrated that few-shot learning-based text classification models are very successful \cite{Holtzman2020}. Theoretically, few-shot learning-based text classification models are possible when the target domain's sample size is insufficient. Second, model generated texts are sampled from a specific distribution and are more regular than human-written texts, and more easily recognized by text classification models \cite{Ippolito2020}. As a result, a few-shot learning-based text classification model can be used as the reward function.

Furthermore, if the reward function fails to accurately determine whether the texts are generated or human written, the RL agent has a tendency to follow a particular generative pattern. In this generative pattern, the agent generates special texts that can trick the reward function into yielding higher reward signals, but these texts are typically unreadable to humans. For this reason, adversarial training is introduced to strengthen the text classification model's robustness and ensure that the agent is appropriately rewarded. RL is combined with a generative adversarial network (GAN) after adversarial training. In the following description, the reward function or text classification model will be referred to as a discriminator, and the agent or generative model will be referred to as a generator. 

\subsection{Active Learning}

During RL, agents are not actively informed of new knowledge about the environment or what the best action is. The environment is the only source of direct instruction for agents, and even then, it only provides rewards rather than actual knowledge. By gaining limited new knowledge directly, humans can quicken their learning. This is especially important for NLP, where new knowledge refers to new words, expressions, or word interpretations (ambiguities in language). Words cannot be generated by memory, so knowledge based on new vocabulary cannot be learned indirectly. Additionally, indirect learning takes a lot of time because agents continuously try out new actions to maximize reward.
 
This study integrates MLE into RL to ensure that agents can directly acquire samples to address this problem. The ability to assign a high probability to the tokens updated by a likelihood is inspired by the work of unlikelihood training \cite{Lagutin}. However, this research differs from unlikelihood training. The aim of unlikelihood training is to improve neural text degeneration \cite{Welleck2019}. To reduce the likelihood of previously generated tokens, it regulated likelihood loss by unlikelihood loss \cite{Lagutin,Welleck2019}. This study aims to provide agents with direct access to samples of the target domain and increase the probability of tokens that are more likely to be used in the target domain.

\subsection{Process of Learning}

Following the work of SeqGAN \cite{Yu2017}, the same policy gradient is adopted and Monte Carlo (MC) search is employed to approximate the state-action value. The objective of the generator model $G_{\theta}(y_{t}|Y_{1:t-1})$ is to generate a text from the start state $s_{0}$ to maximize its reward:

\begin{equation}
	J(\theta)=\mathbb{E}[R_{T}|s_{0},\theta]=\sum_{y\in Y}G_{\theta}(y|s_{0}) \cdot Q_{D_{\phi}}^{G_{\theta}}(s_{0},y),
	\label{eq.objective}
\end{equation}

\noindent where $Q_{D_{\phi}}^{G_{\theta}}(s,a)$ is the action-value function of a text. $R_{T}$ is the reward for a complete sequence and the reward from discriminator $D_{\phi}$. 

However, the discriminator only provides a reward value for a finished sequence and every action should be rewarded. Thus, MC search with generator policy $G_{\theta}$ is applied to sample the unknown last tokens. $N$-time MC search be represented as

\begin{equation}
	\{Y_{1:T}^{1},...,Y_{1:T}^{N}\}=MC^{G_{\theta}}(Y_{1:t};N),
	\label{eq.MCMC}
\end{equation}

\noindent where $Y_{1:t}^{n}$ and $Y_{t+1:T}^{n}$ is sampled based on roll-out policy $G_{\theta}$ and the current state. Thus, it has:

\begin{equation}
	Q_{D_{\phi}}^{G_{\theta}}(a=y_{t},s=Y_{1:t-1})=\left \{
	\begin{array}{lcr}
		\frac{1}{N}\sum_{n=1}^{N}D_{\phi}(Y_{1:T}^{n}) & for \quad  t<T\\
		D_{\phi}(Y_{1:t}) & for \quad t=T,
	\end{array}
	\right.
	\label{eq.action-value}
\end{equation}

\noindent where $Y_{1:T}^{n} \in MC^{G_{\theta}}(Y_{1:t};N)$.

The generator's parameters are updated in RL as:

\begin{equation}
	\theta \leftarrow \theta+\alpha \nabla_{\theta}J(\theta),
	\label{eq.theta_update}
\end{equation}

\noindent where $\alpha$ is the learning rate. And $\nabla_{\theta}J(\theta)$ is an unbiased estimation as:

\begin{equation}
	\nabla_{\theta}J(\theta)=\frac{1}{T}\sum_{t=1}^{T}\mathbb{E}_{y_{t}\sim G_{\theta}(y_{t}|Y_{1:t-1})}[\nabla_{\theta}\log G_{\theta}(y_{t}|Y_{1:t-1}) \cdot Q_{D_{\phi}}^{G_{\theta}}(Y_{1:t-1},y_{t})].
	\label{eq.estimation}
\end{equation}

\begin{algorithm}[h]
	\caption{The training process in DARL}
	\small
	\label{alg.train}
	\begin{algorithmic}[1]
		\Require generator policy $G_{\theta}$; discriminator $D_{\phi}$
		\Require update rate $R$
		\Require a source dataset $S=\{X_{1:m}^{1},...,X_{1:m}^{N}\}$ includes $N$ domains; a target domain dataset $S^{'}=\{X_{1:k}\}$
		
		\State Initialize $G_{\theta}$ and $D_{\phi}$ with random weights $\theta$ and $\phi$
		\State Pre-train $G_{\theta}$ using MLE on $S$
		\State Pre-train $D_{\phi}$ using few-shot learning on $S$
		\State Generate negative samples using $G_{\theta}$ for training $D_{\phi}$
		\State train $D_{\phi}$ using few-shot learning on negative samples and $S^{'}$  
		
		\While{not done}
			\State sample $r \sim U[0,1]$
			\If{$r<R$}
				\State Generate a sequence $Y_{1:T}=(y_{1},...,y_{T}) \sim G_{\theta}$
				\For{$t$ in $1:T$}
					\State Compute $Q(a=y_{t};S=Y_{1:t-1})$ by Eq.\ref{eq.action-value}
				\EndFor 
				\State Update generator parameters via a policy gradient Eq.\ref{eq.theta_update}
			\Else
				\State Update generator parameters via MLE on $S^{'}$
			\EndIf
			\State Use current $G_{\theta}$ to generate negative examples and combine with $S^{'}$
			\State Train discriminator $D_{\phi}$ by few-shot learning
		\EndWhile
	\end{algorithmic}
\end{algorithm}

In summary, algorithm \ref{alg.train} shows full details of the proposed DARL. At the beginning of the training, $G_{\theta}$ is pre-trained by MLE on source dataset $S$. Despite containing $N$ domains, $S$ is considered as a whole. The category information from the domain is ignored during $G_{\theta}$'s pre-training. And $D_{\phi}$ is also pre-trained using few-shot learning on $S$. Each domain in $S$ represents a specific task. The category information for the domain is taken into account during $D_{\phi}$'s pre-training.

After pre-training, the generator and discriminator are trained alternatively. The ratio of RL training to MLE training is determined by the hyperparameter $R$, and the generator is randomly trained by either method. The discriminator needs to be trained periodically to keep up with the generator as it advances through training. Positive examples for the discriminator's training come from the provided dataset $S^{'}$, whereas negative examples come from $G_{\theta}$. And the number of negative examples is the same as the number of positive examples.

The RL training of DARL is similar to SeqGAN. However, they are fundamentally distinct. SeqGAN adopts RL and MC search to generate sequences of discrete tokens because GANs are designed for generating real valued and continuous data. SeqGAN extents the application fields of GANs. DARL adopts RL and MC search to find phrases with target domain features and increase the likelihood of certain tokens. And it adopts the adversarial ideal to improve robustness of the reward function. DARL achieves domain adaptation for text generation using RL.

\section{Experiment}

The proposed framework was evaluated by conducting experiments on the dataset. According to the common experiment configuration of few-shot learning, a 5-shot learning dataset and a 10-shot learning dataset were created, respectively.

\subsection{Baselines}

To evaluate its effectiveness, DARL was compared with the following five baselines: 

\begin{itemize}
	\item[$\bullet$] 
	\textbf{Fine-tune} fine-tunes the pre-trained base model on the support sets of each target domain. It can be viewed as a special case of DARL where the hyperparameter $R$ of DARL is set to 0 and no RL is done.
	\item[$\bullet$] 
	\textbf{MetaNLG} \cite{Mi2019} is a generalized optimization-based approach based on the well-recognized MAML algorithm. It learns a better initialization of model parameters that facilitates fast adaptation to new low-resource domains.
	\item[$\bullet$] 
	\textbf{DAML} \cite{Qian2020} is an end-to-end trainable generation model based on meta-learning that learns from multiple rich-resource tasks and then adapts to new domains. It combines multiple tasks in training to learn general and transferable information that is applicable to new domains.
	\item[$\bullet$] 
	\textbf{MemIML} \cite{Zhao2022} is a memory imitation meta-learning method that enhances the model’s reliance on support sets for task adaptation. It introduces a task-specific memory module to address memorization overfitting issue.
	\item[$\bullet$] 
	\textbf{SeqGAN} \cite{Yu2017} is a sequence generation framework based on GAN to generate sequences of discrete tokens. It can be viewed as a special case of DARL where the hyperparameter $R$ is set to 1 and no MLE is done.
\end{itemize}

The benchmark was measured using fine-tuning; the advantages and disadvantages of meta-learning were demonstrated using three meta-learning-based baselines; the effectiveness of DRAL for RL improvement was illustrated using a GAN-based baseline.

\subsection{Datasets}

The dataset must include several domains because the meta-learning-based baselines need to sample across a family of domains. The following dataset, which is frequently used in text generation tasks, was used to satisfy this requirement.

\textbf{Amazon Reviews} \cite{McAuley2015} contains product reviews and metadata from Amazon, including 24 products reviews. Five products were picked randomly as source domains, and the other 5 products were chosen randomly as target domains. The sentences with words between 15 and 30 were selected randomly from each domain, resulting in a dataset of 2000 sentences per domain. Toys and games (toys), pet supplies (pet), beauty, grocery and gourmet food (food), and baby were included by source domains. Cell phones and accessories (phones), tools and home improvement (tools), office products (office), automotive and digital music (music) were  included by target domains. In the $k$-shot configuration, $k$ samples of the per target domain were selected randomly, and all methods shared same samples.  

\subsection{Experiment Setup}

A RNN was set as a single-layer LSTM with a hidden dimension size of
256 and a maximum length of 30 words. Word embeddings were randomly initialized to a dimension of 300, and the $R$ of DARL was set to 0.5. The induction networks \cite{Geng} which were proposed for few-shot learning text classification was selected as discriminator. DARL was pre-trained for 100 epochs and trained for 150 epochs. And it was implemented based on Pytorch\footnote{The repeatable experiment code is made publicly available on https://github.com/cps11/DARL.}. 

To account for the instability of few-shot learning, each method was tested ten times, and the model that resulted from each test was sampled ten times with 5000 samples per domain. The mean of the aforementioned results was used to calculate the final experimental results. 

\subsection{Quality of Generated Texts}
\label{sec_quality}

Bilingual evaluation understudy (BLEU) \cite{Papineni2002} and Self-BLEU \cite{Alihosseini2019} were adopted to measure the quality of the generated texts in terms of fluency and diversity.

\begin{itemize}
	\item[$\bullet$] 
	\textbf{BLEU} uses source domains set as a reference and evaluates each generated text using the BLEU 5-gram score. It measures the fluency of the generated texts.	
	\item[$\bullet$] 
	\textbf{Self-BLEU} uses generated texts as a reference and evaluates each generated text using Self-BLEU 5-gram score. It measures the diversity of the generated texts.	
\end{itemize}

The results are displayed in Table \ref{tab.quality.5shot} and Table \ref{tab.quality.10shot}. Comparing with baselines,  DARL improves not only the diversity of the generated text (lower Self-BLEU), but also the fluency of the generated text (higher BLEU). 

Some methods used to generate texts had a high degree of similarity to samples when they were manually examined. To quantify this phenomenon, Eq.\ref{eq.similarity} was used to identify the sample with which each generated text was most similar. And the similarity was evaluated using $BLEU(s^{*},y)$. 

\begin{equation}
	s^{*}=\arg\max_{s\in S^{'}}BLEU(s,y), y\in Y
	\label{eq.similarity}
\end{equation}

\begin{table}[H]
	\centering
	\caption{Quality of the generated texts on the 5-shot dataset}
	\label{tab.quality.5shot}
	\resizebox{\textwidth}{!}{
		\begin{tabular}{cccccccc}
			\toprule
			Metrics & Methods & Automotive & Music & Office & Phone & Tools & Average\\ 
			\midrule
			\multirow{6}*{BLEU$\uparrow$} & Fine-tune & 0.328 & 0.263 & 0.376 & 0.446 & 0.407 & 0.364\\
			~ & MetaNLG & 0.354 & 0.275 & 0.386 & 0.459 & 0.421 & 0.379\\
			~ & DAML & 0.390 & 0.273 & 0.400 & 0.439 & \textbf{0.433} & 0.387\\
			~ & MemIML & 0.136 & 0.112 & 0.170 & 0.117 & 0.123 & 0.132\\	
			~ & SeqGAN & 0.199 & 0.142 & 0.302 & 0.258 & 0.313 & 0.243\\
			~ & \textbf{DARL} & \textbf{0.391} & \textbf{0.340} & \textbf{0.418} & \textbf{0.472} & 0.405 & \textbf{0.405}\\	
			\hline	
			\multirow{6}*{Self-BLEU$\downarrow$} & Fine-tune & 0.932 & 0.932 & 0.943 & 0.949 & 0.946 & 0.940\\
			~ & MetaNLG & 0.852 & 0.829 & 0.866 & 0.935 & 0.927 & 0.882\\
			~ & DAML & 0.902 & 0.825 & 0.896 & 0.947 & 0.938 & 0.902\\
			~ & MemIML & 0.995 & 0.993 & 0.995 & 0.994 & 0.994 & 0.994\\	
			~ & SeqGAN & 0.813 & 0.899 & 0.936 & 0.910 & 0.949 & 0.901\\
			~ & \textbf{DARL} & \textbf{0.812} & \textbf{0.822} & \textbf{0.842} & \textbf{0.875} & \textbf{0.860} & \textbf{0.842}\\	
			\bottomrule
		\end{tabular}
	}
\end{table}

\begin{table}[H]
	\centering
	\caption{Quality of the generated texts on the 10-shot dataset}
	\label{tab.quality.10shot}
	\resizebox{\textwidth}{!}{
		\begin{tabular}{cccccccc}
			\toprule
			Metrics & Methods & Automotive & Music & Office & Phone & Tools & Average\\ 
			\midrule
			\multirow{6}*{BLEU$\uparrow$} & Fine-tune & 0.341 & 0.324 & 0.418 & 0.322 & 0.355 & 0.352\\
			~ & MetaNLG & 0.326 & 0.300 & 0.419 & 0.307 & 0.347 & 0.340\\
			~ & DAML & 0.314 & 0.302 & \textbf{0.429} & 0.314 & 0.351 & 0.342\\
			~ & MemIML & 0.132 & 0.133 & 0.077 & 0.147 & 0.067 & 0.111\\	
			~ & SeqGAN & 0.197 & 0.278 & 0.249 & 0.199 & 0.233 & 0.231\\
			~ & \textbf{DARL} & \textbf{0.378} & \textbf{0.397} & 0.419 & \textbf{0.365} & \textbf{0.393} & \textbf{0.391}\\	
			\hline	
			\multirow{6}*{Self-BLEU$\downarrow$} & Fine-tune & 0.865 & 0.875 & 0.888 & 0.863 & 0.873 & 0.873\\
			~ & MetaNLG & 0.789 & 0.767 & 0.849 & 0.868 & 0.805 & 0.815\\
			~ & DAML & 0.843 & 0.854 & 0.896 & 0.897 & 0.874 & 0.873\\
			~ & MemIML & 0.992 & 0.994 & 0.995 & 0.993 & 0.992 & 0.993\\	
			~ & SeqGAN & 0.882 & 0.905 & 0.901 & 0.877 & 0.944 & 0.902\\
			~ & \textbf{DARL} & \textbf{0.689} & \textbf{0.691} & \textbf{0.726} & \textbf{0.673} & \textbf{0.737} & \textbf{0.703}\\	
			\bottomrule
		\end{tabular}
	}
\end{table}

\begin{figure}[H]
	\centering
	\subfigure[automotive]{
		\includegraphics[width=0.3\textwidth]{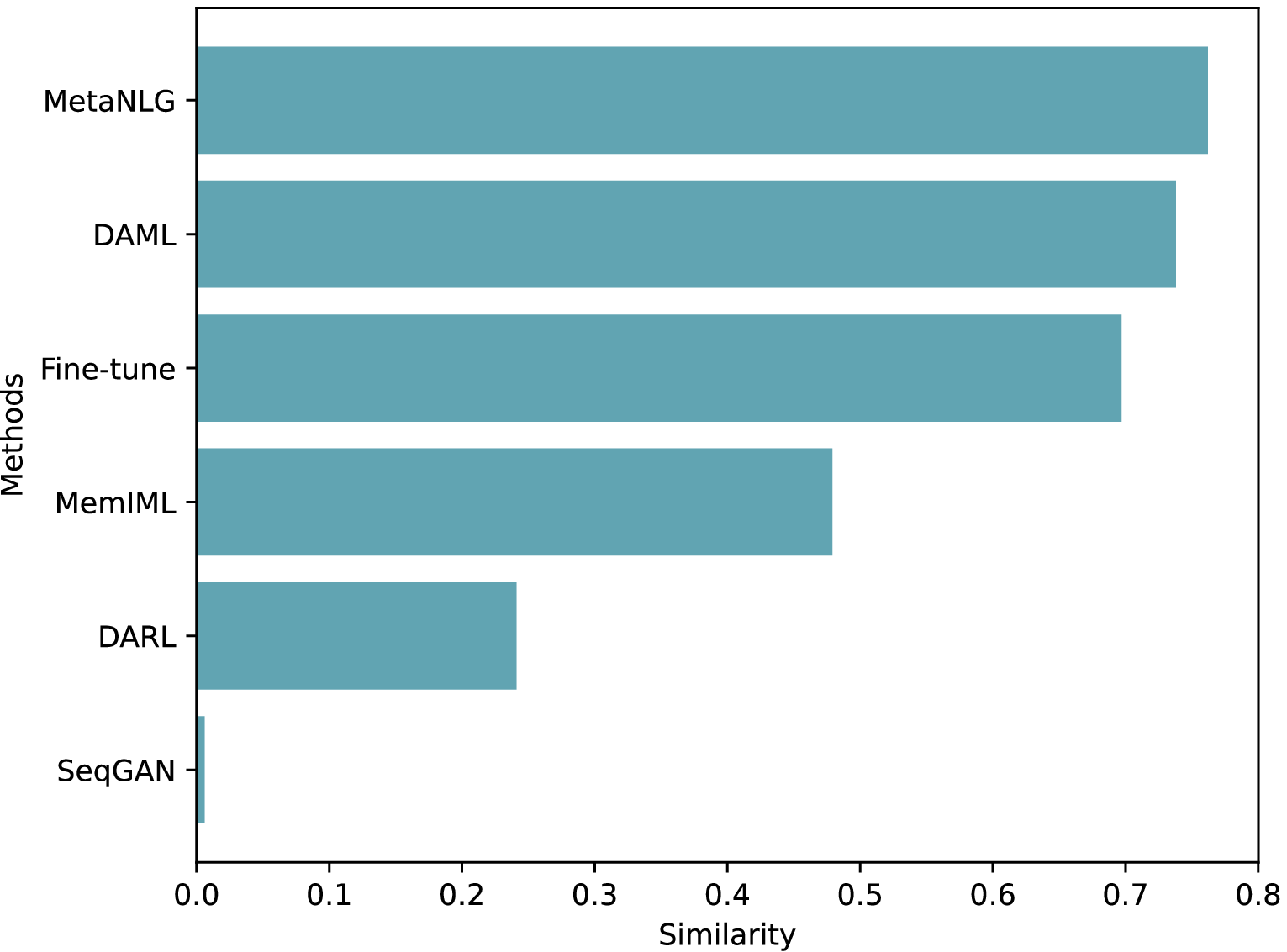}
	}
	\subfigure[music]{
		\includegraphics[width=0.3\textwidth]{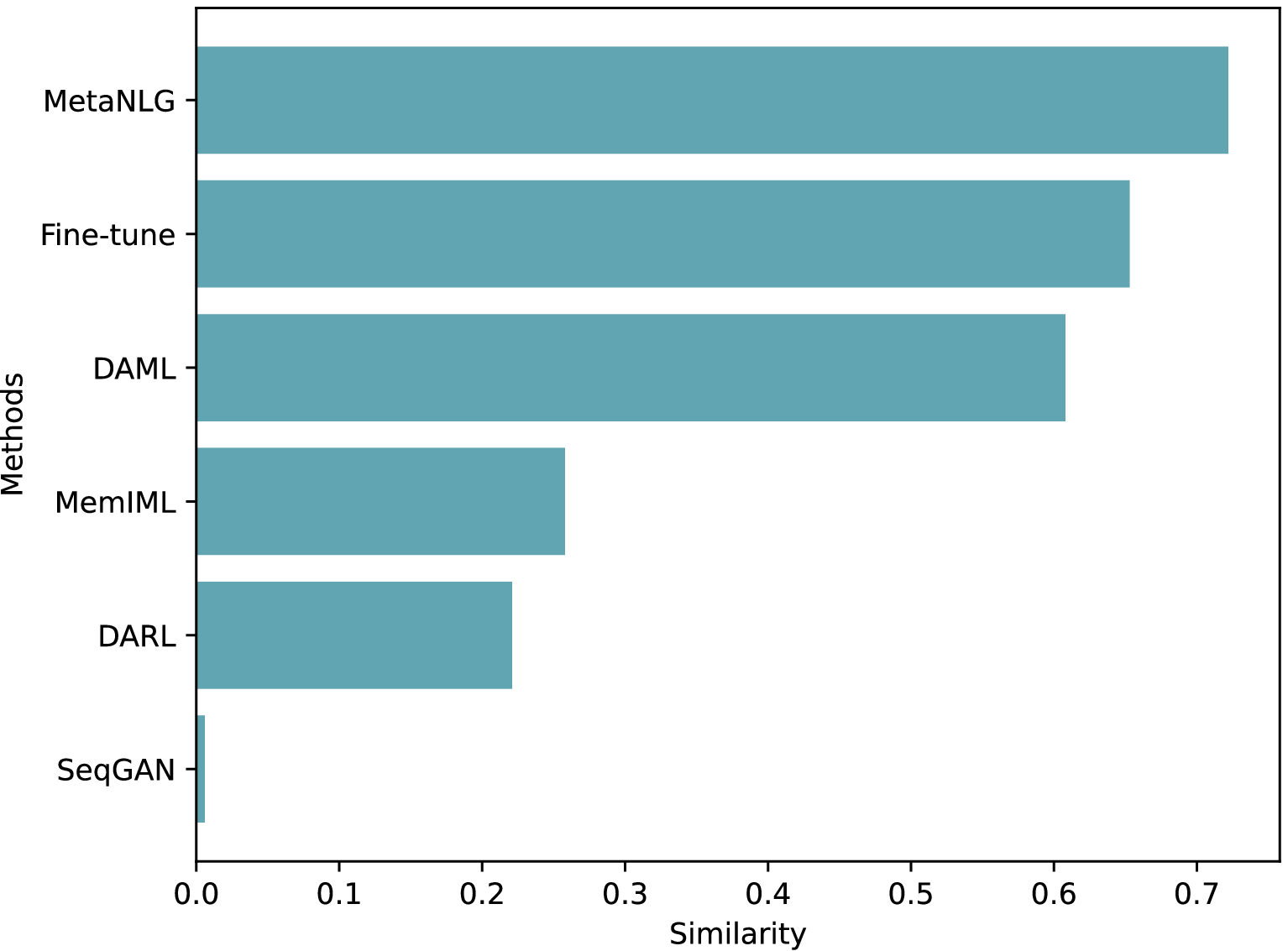}
	}
	\subfigure[office]{
		\includegraphics[width=0.3\textwidth]{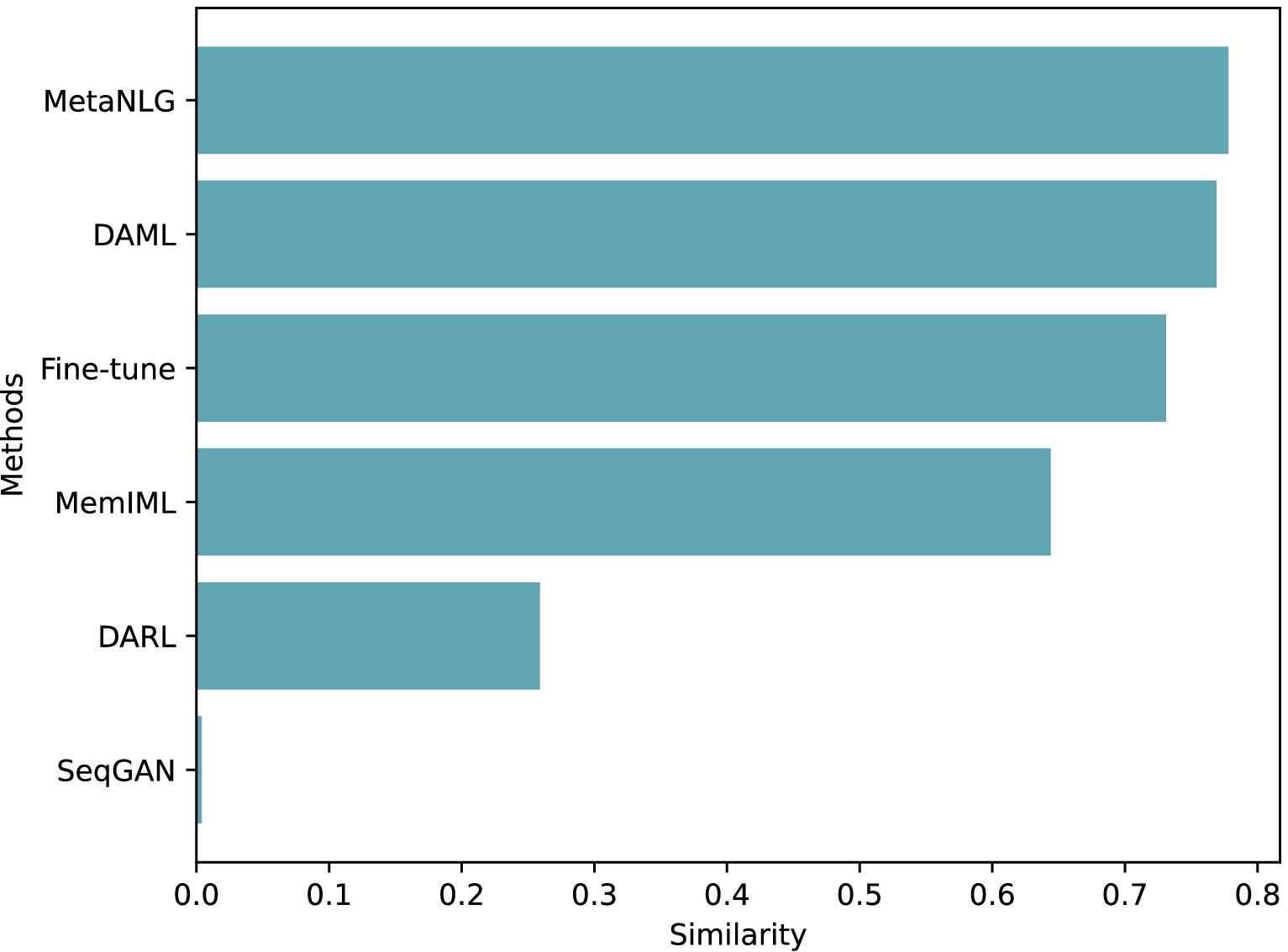}
	}
	\subfigure[phone]{
		\includegraphics[width=0.3\textwidth]{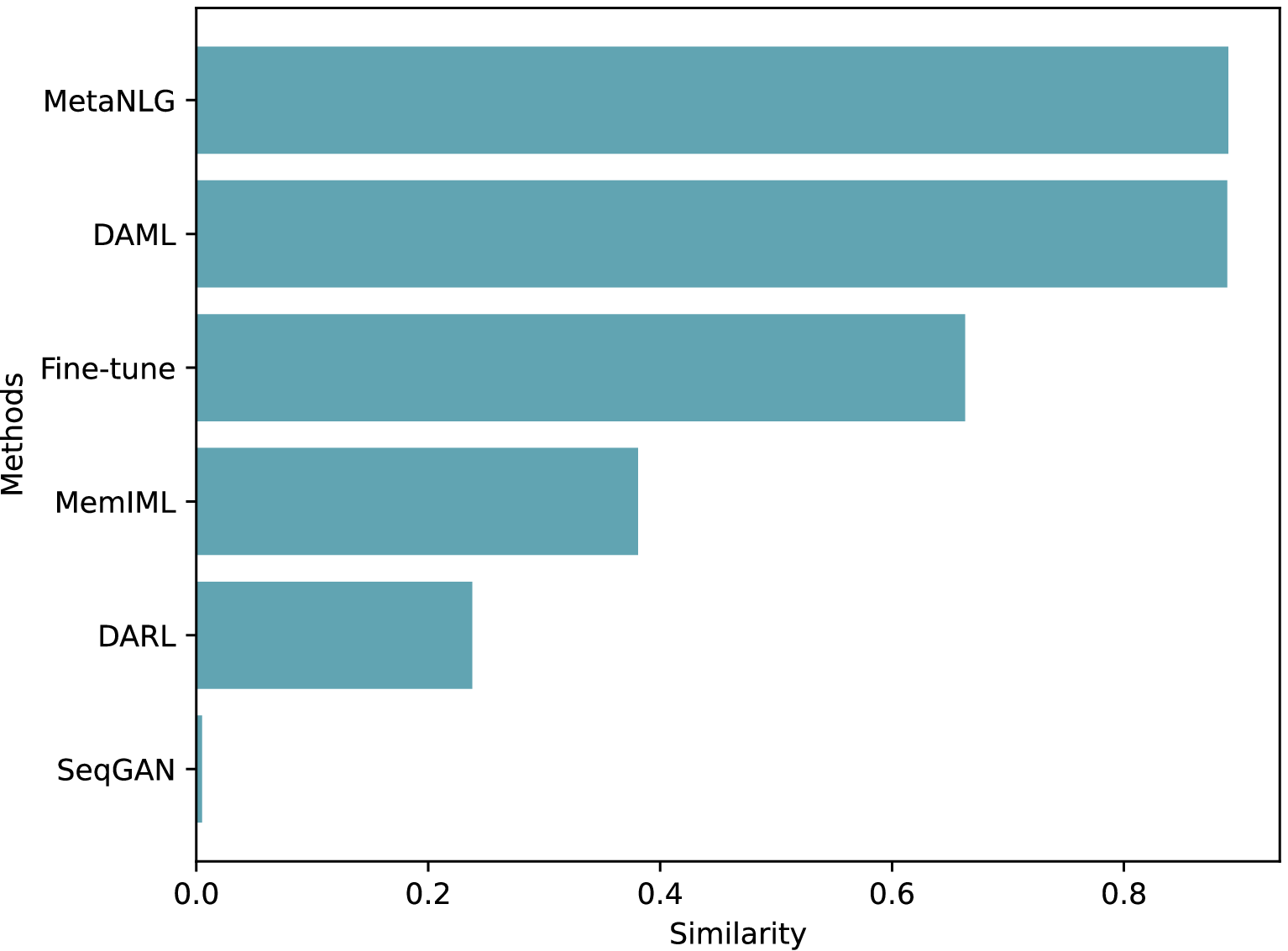}
	}
	\subfigure[tools]{
		\includegraphics[width=0.3\textwidth]{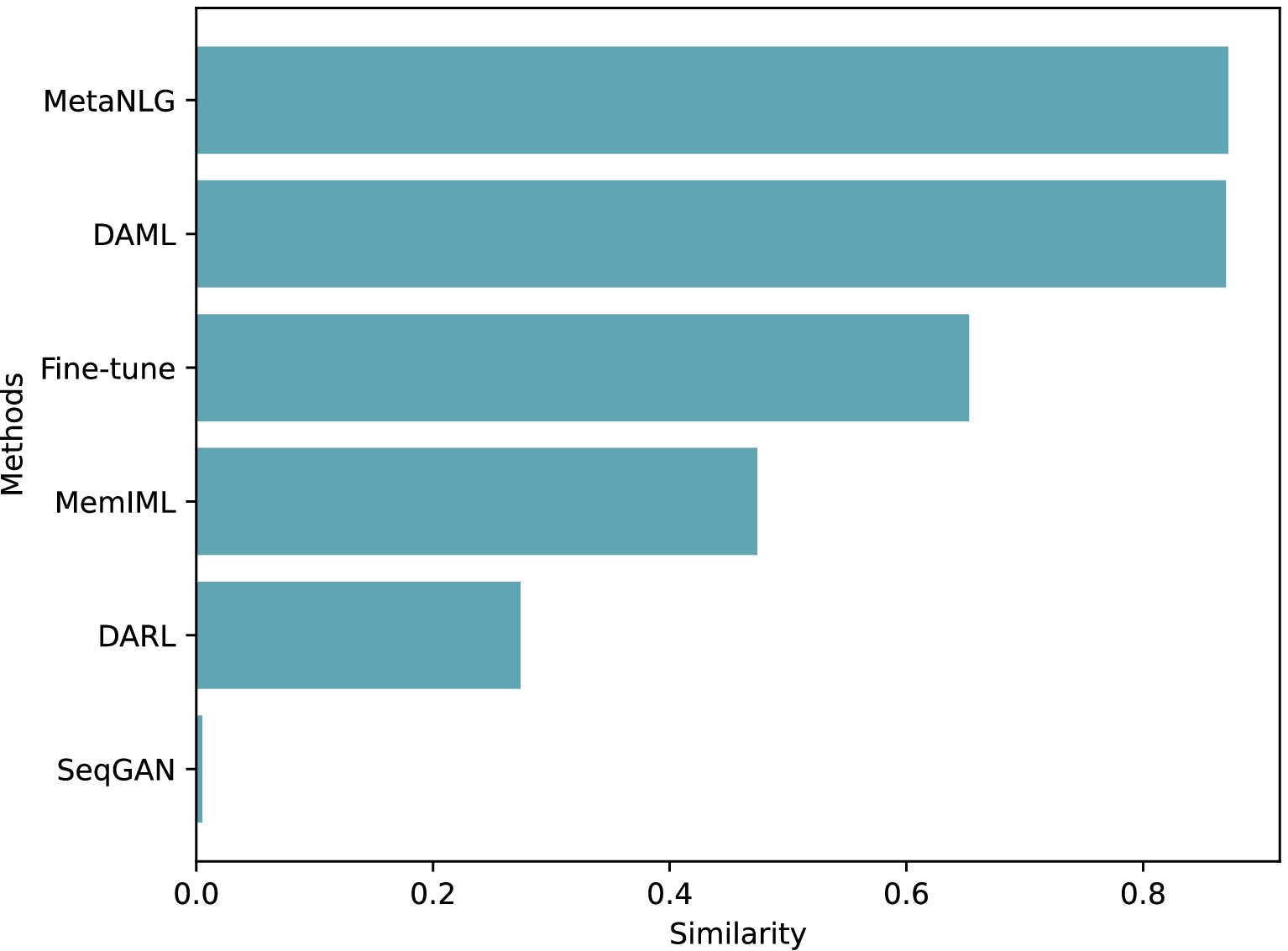}
	}
	\subfigure[average]{
		\includegraphics[width=0.3\textwidth]{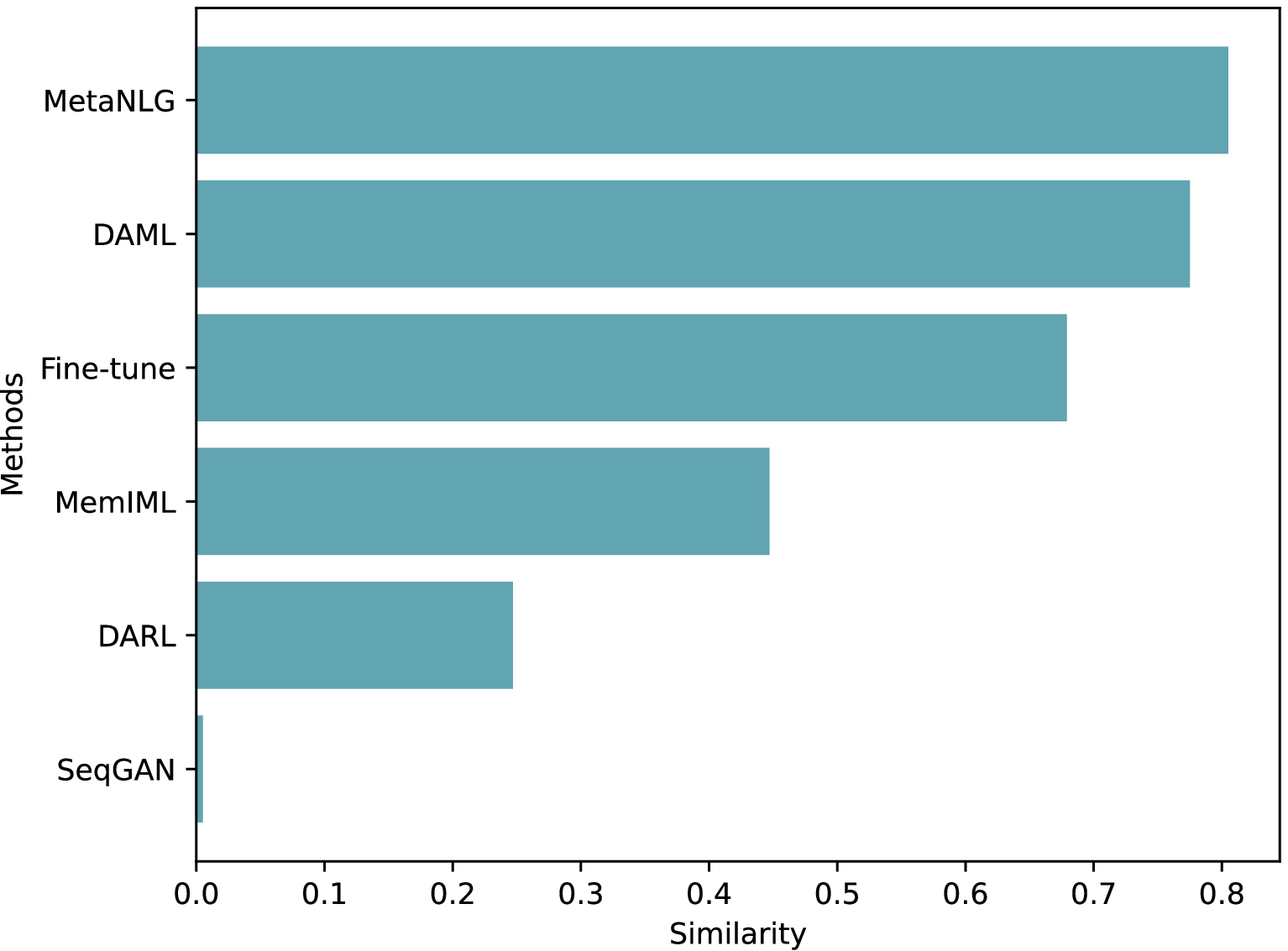}
	}
	\caption{The similarity between generated texts and samples on the 5-shot database. As the bar is raised, generated texts and samples become more similar. The mean value of results across all target domains is shown in the subfigure titled average.}
	\label{fig.similarity.5shot}
\end{figure}

\begin{figure}[t]
	\centering
	\subfigure[automotive]{
		\includegraphics[width=0.3\textwidth]{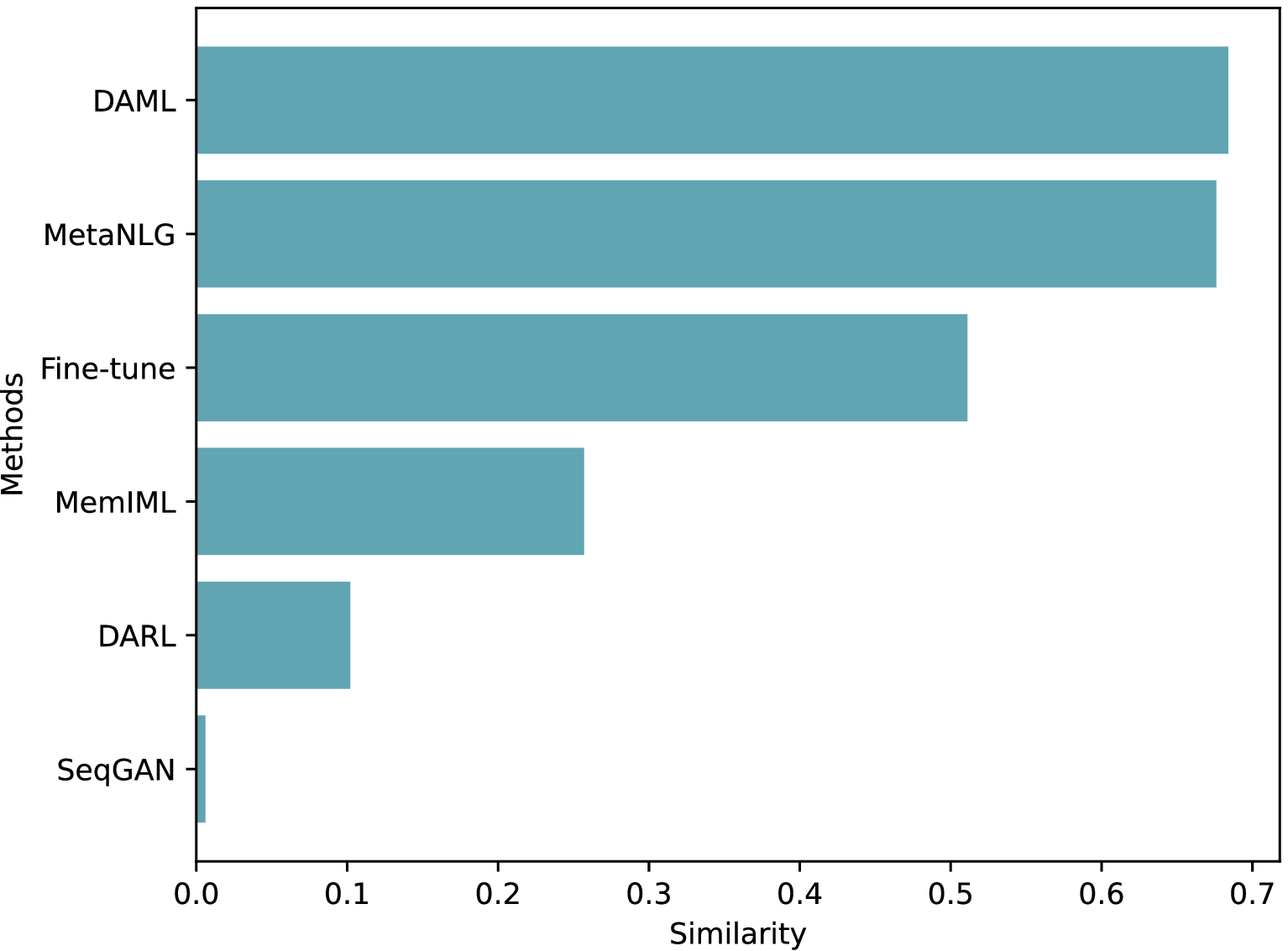}
	}
	\subfigure[music]{
		\includegraphics[width=0.3\textwidth]{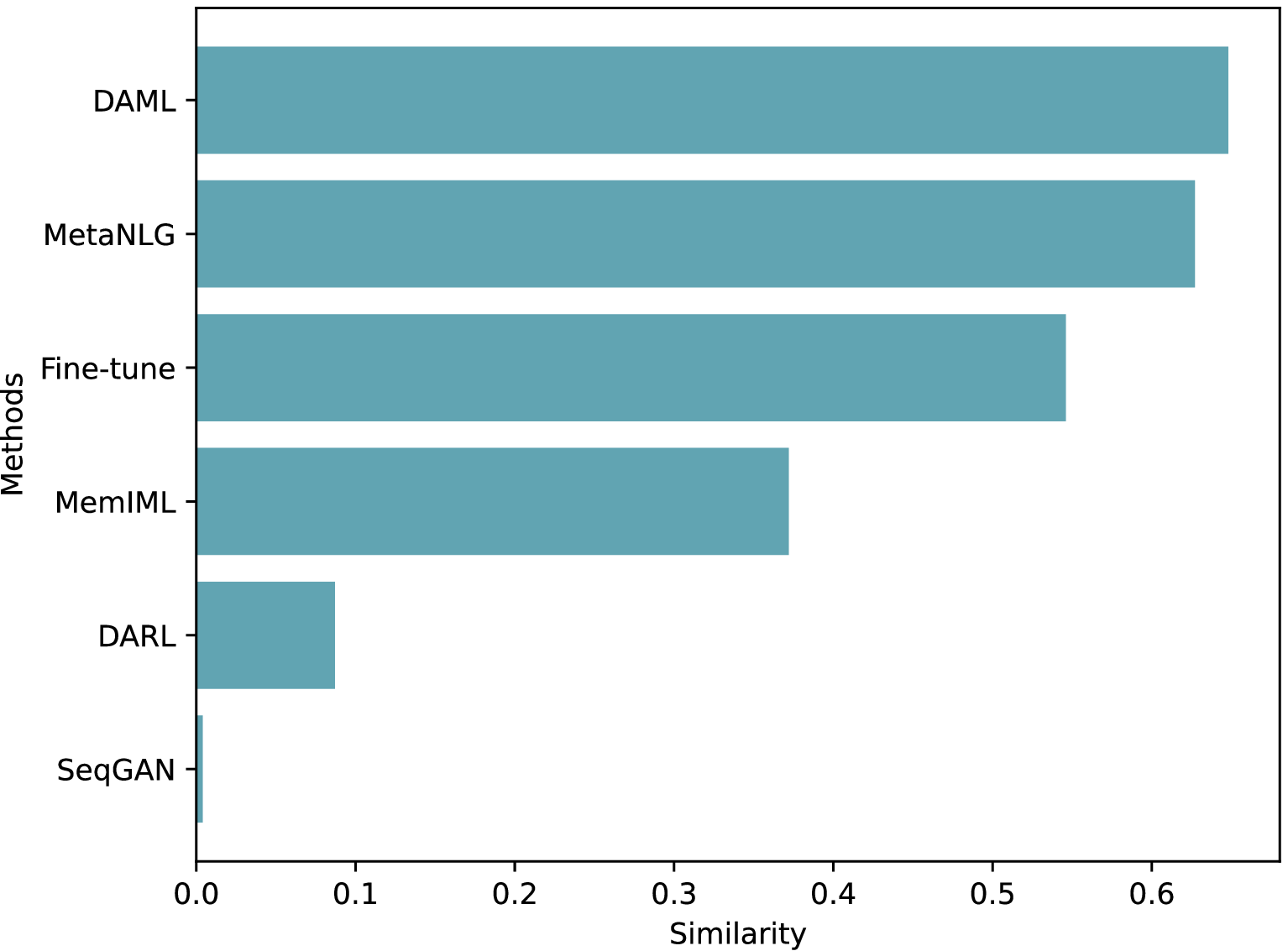}
	}
	\subfigure[office]{
		\includegraphics[width=0.3\textwidth]{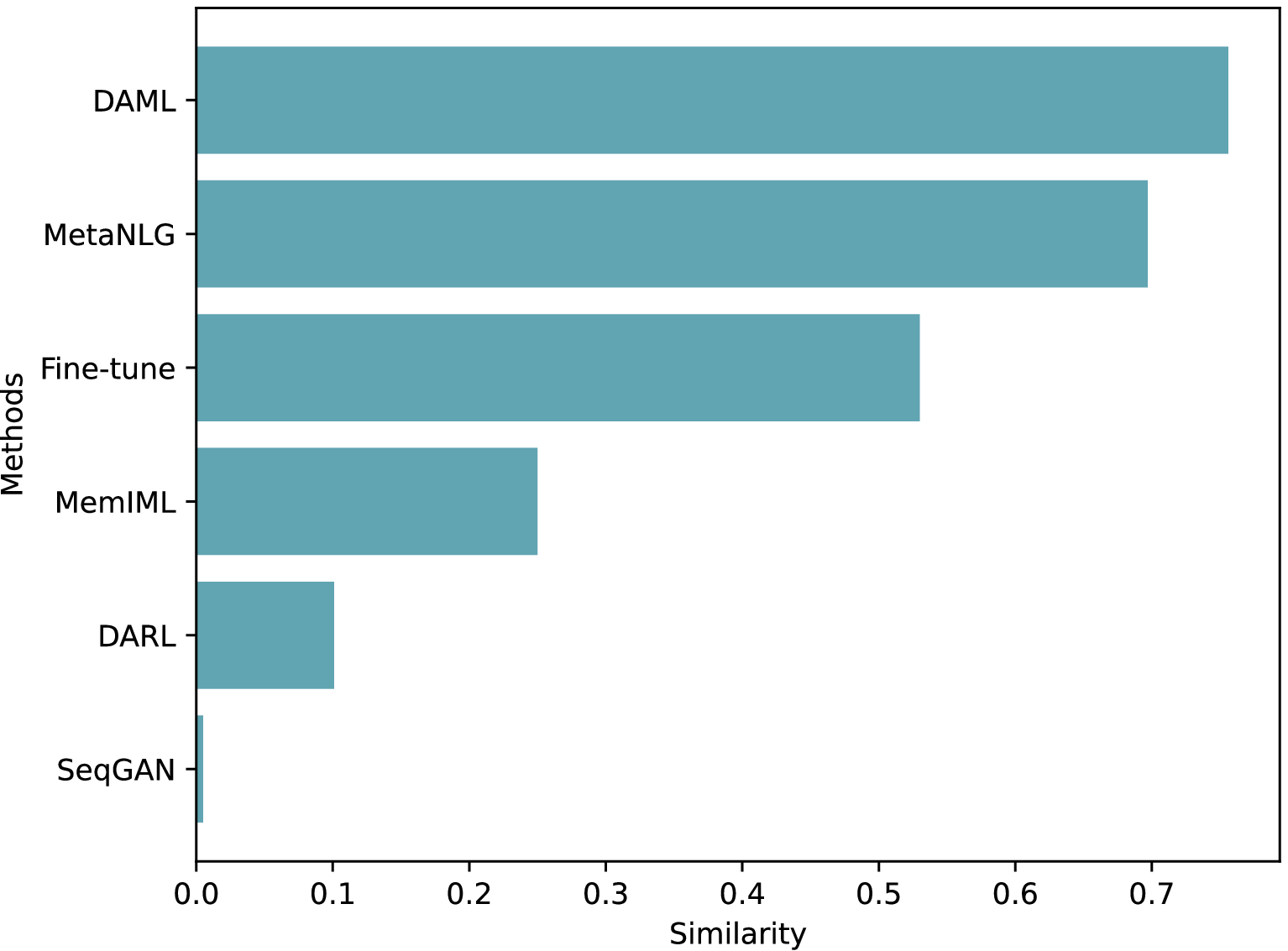}
	}
	\subfigure[phone]{
		\includegraphics[width=0.3\textwidth]{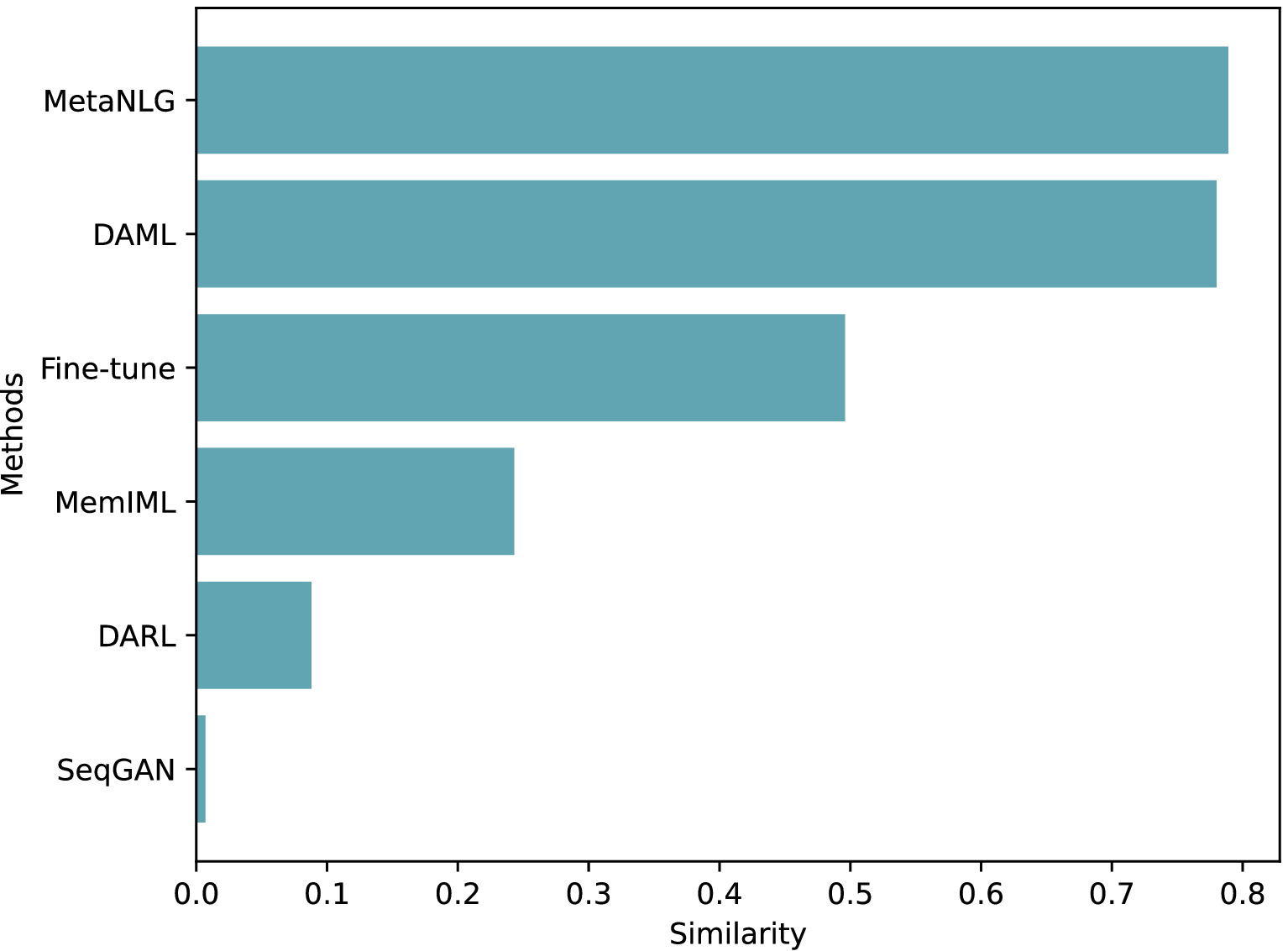}
	}
	\subfigure[tools]{
		\includegraphics[width=0.3\textwidth]{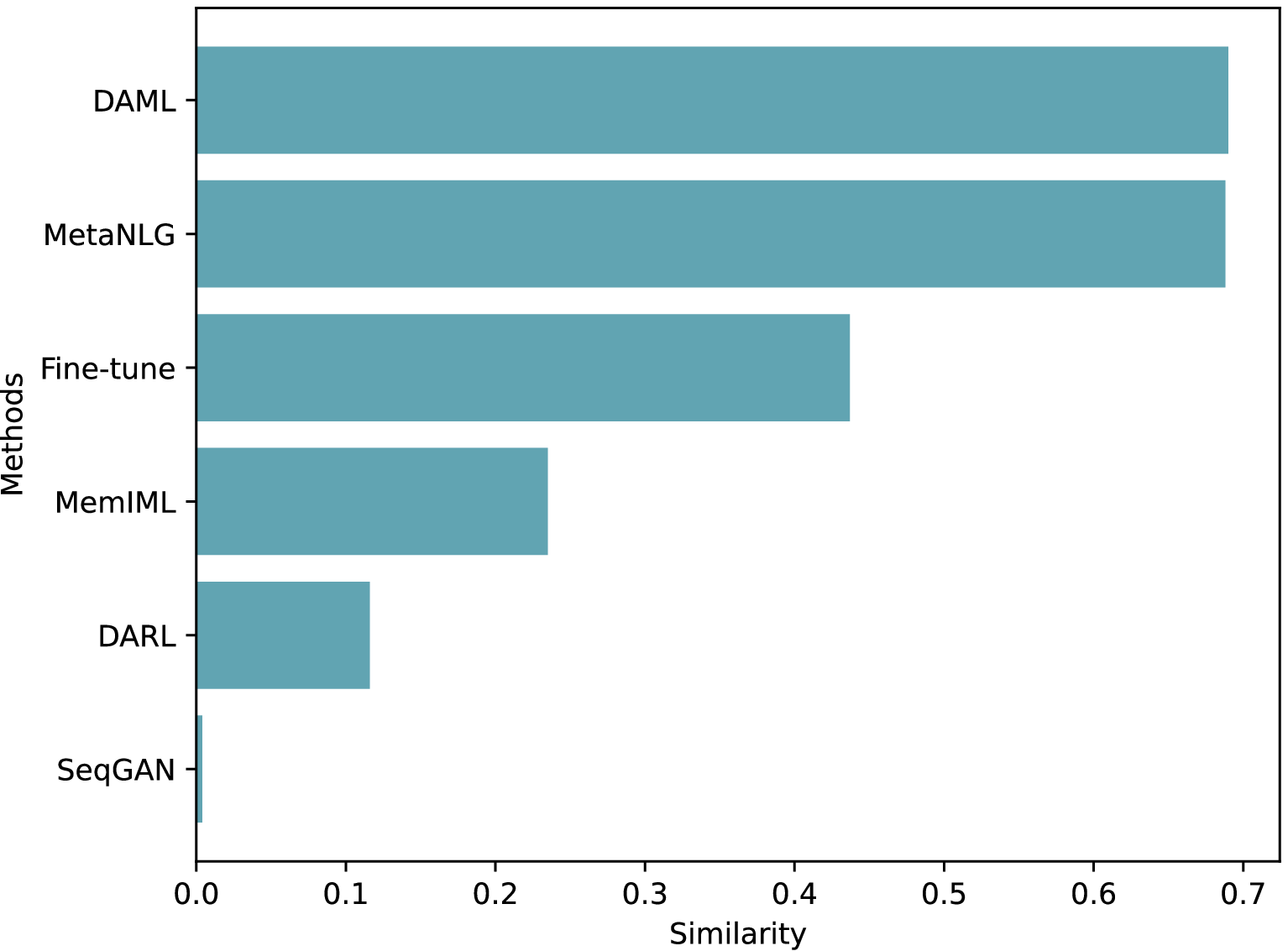}
	}
	\subfigure[average]{
		\includegraphics[width=0.3\textwidth]{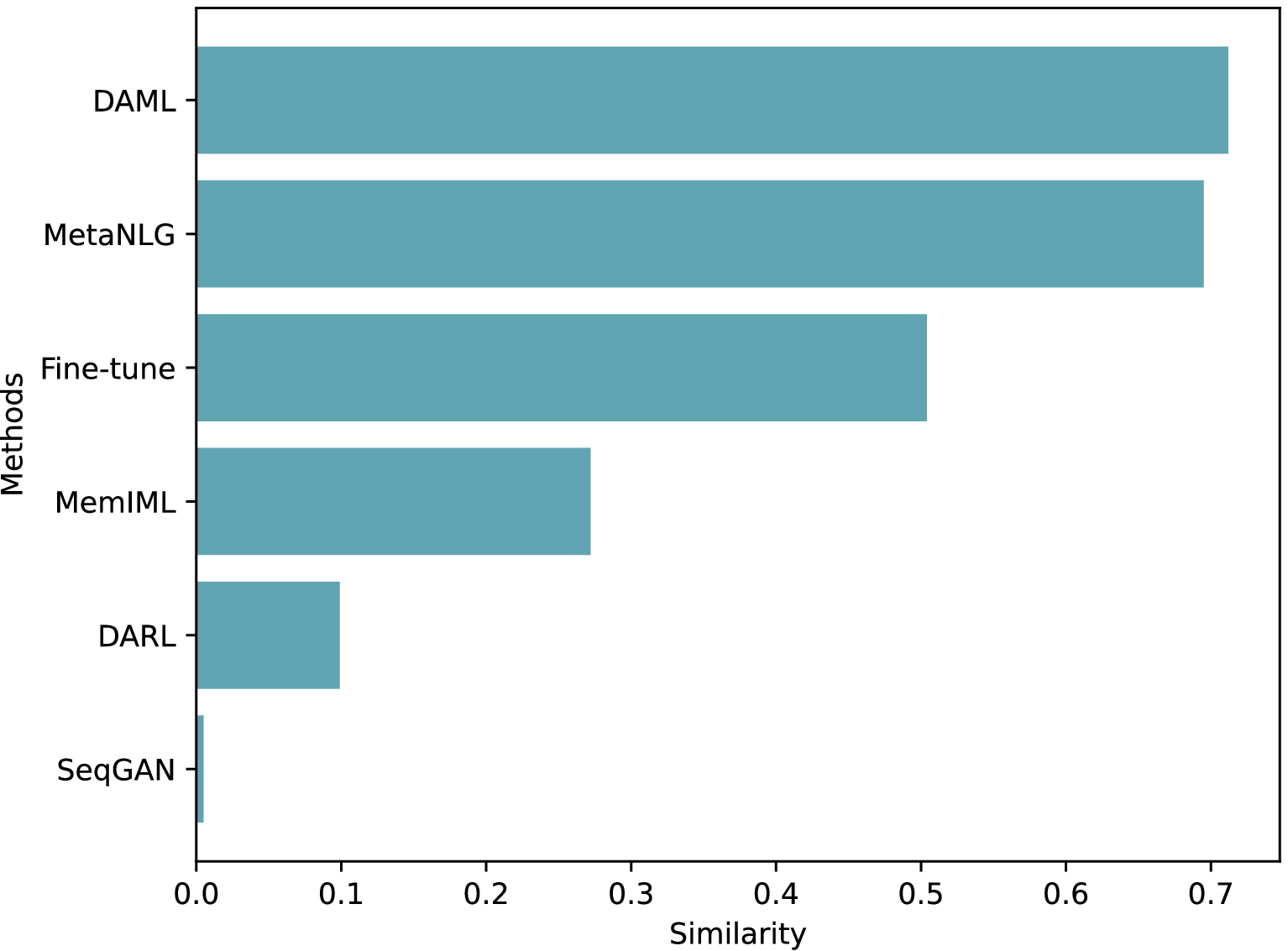}
	}
	\caption{The similarity between generated texts and samples on the 10-shot database. As the bar is raised, generated texts and samples become more similar. The mean value of results across all target domains is shown in the subfigure titled average.}
	\label{fig.similarity.10shot}
\end{figure}

As shown in Figure \ref{fig.similarity.5shot} and Figure \ref{fig.similarity.10shot}, the methods can be split into two parts using Fine-tune method as a benchmark. METANLG and DAML, which are based on meta-learning, have higher similarity than Fine-tune method. This suggests that while these two methods are more capable of learning, they are also more prone to overfitting. MemIML, DARL and SeqGAN have lower similarity than Fine-tune method. The main objective of MemIML, which is also based on meta-learning, is to address the issue of memorization overfitting in meta-learning. Although MemIML's similarity is lower than that of the other two meta-learning-based methods, it is still higher than that of DARL and SeqGAN. SeqGAN achieves the lowest similarity because it is not directly trained on samples. Its higher Self-BLEU value, however, suggests that a pattern collapse had taken place. Pattern collapse causes generated texts to follow only a few fixed patterns, which results in a high degree of similarity between generated texts even though they differ from samples. Because DARL can be trained directly with samples through MLE, the similarity between its generated texts and samples is higher than that between SeqGAN generated texts and samples. However, the phenomenon that the similarity between texts generated by DARL is lower than the similarity between texts generated by SeqGAN illustrates that adding MLE to RL can improve the problem of GAN pattern collapse.

The analysis that was just mentioned explains the rationale for each method's diversity indicator (Self-BLEU). Texts generated based on meta-learning methods are similar to each other because they are highly similar to samples. Texts generated by these methods are less diverse. The Fine-tune method is the same as the meta-learning based methods. Due to pattern collapse, texts generated by SeqGAN have a few fixed patterns. They are also similar to each other and less diverse. The diversity of the texts generated by DARL outperforms the baselines because they are less similar to the samples or do not experience to pattern collapse like SeqGAN.

In conclusion, the fluency and diversity of the texts generated by DARL are higher than the baselines. In the meantime, the pattern collapse issue can be improved by implementing MLE in RL.

\subsection{Domain relevance of the generated text}

The accuracy of text classification was adopted to measure the domain relevance of generated texts. A state-of-art text classifier \cite{Kim2014} based on convolutional neural network (CNN) was trained independently to automatically evaluate the semantic relevance of generated texts. And a larger dataset than the one utilized for the generation task was used to train the classifier. 

As mentioned in Table \ref{tab.quality.5shot} and Table \ref{tab.quality.10shot}, all baselines have higher Self-BLEU metrics which measure repetition within generated texts. Therefore, some models generate texts that will contain exactly the same texts. The accuracy obtained by measuring the generated texts containing duplicate texts is falsely high. When there is a high similarity between a generated text and the sample, more duplicate texts imply an increased likelihood of similarity between generated texts and samples. As a result, it is unfair to compare various methods based on inflated results.

To objectively and fairly reflect the performance of various methods, concepts of gross profit margin and net profit margin in business accounting are introduced to define the following concepts.

\begin{itemize}
	\item[$\bullet$] 
	\textbf{Sales} are the total amount of generated texts.
	\item[$\bullet$]
	\textbf{Cost} is the number of texts generated in non-target domain as determined by the classifier.
	\item[$\bullet$]
	\textbf{Tax} is the number of generated texts that are duplicated after non-target domain texts are excluded.
\end{itemize}

Gross accuracy and net accuracy were calculated using Eq.\ref{eq.gross} and Eq.\ref{eq.net}, respectively. The duplicate generated texts are treated as a tax penalty in Eq.\ref{eq.net}, just like the net profit margin is. The tax should be higher as a penalty because generating more duplicate texts results in more waste.

\begin{equation}
	Gross \ Accuracy = \frac{Sales - Cost}{Sales}
	\label{eq.gross}
\end{equation}

\begin{equation}
	Net \ Accuracy = \frac{Sales - Cost - Tax}{Sales}
	\label{eq.net}
\end{equation}

\begin{figure}[H]
	\centering
	\subfigure[automotive]{
		\includegraphics[width=0.3\textwidth]{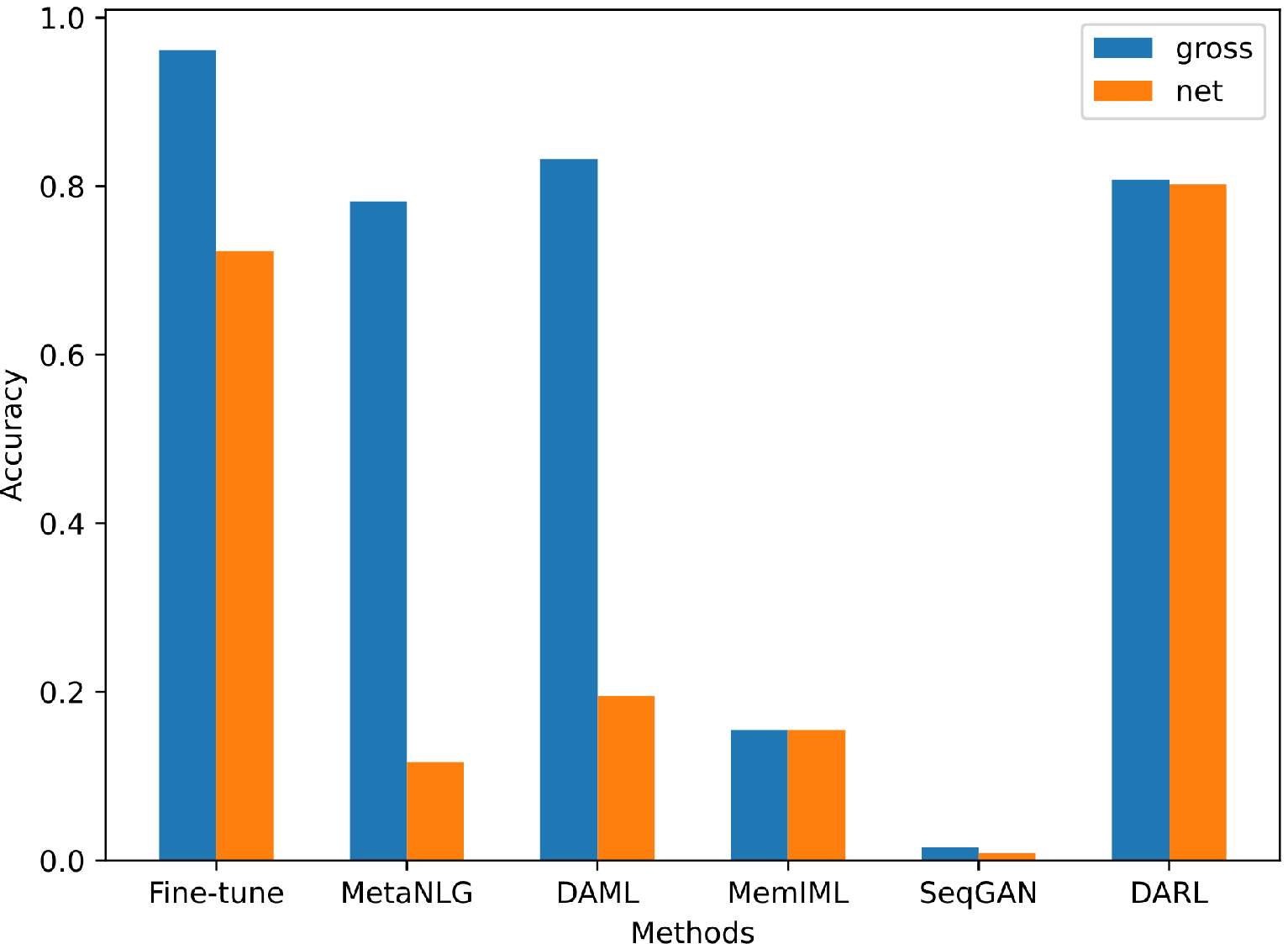}
	}
	\subfigure[music]{
		\includegraphics[width=0.3\textwidth]{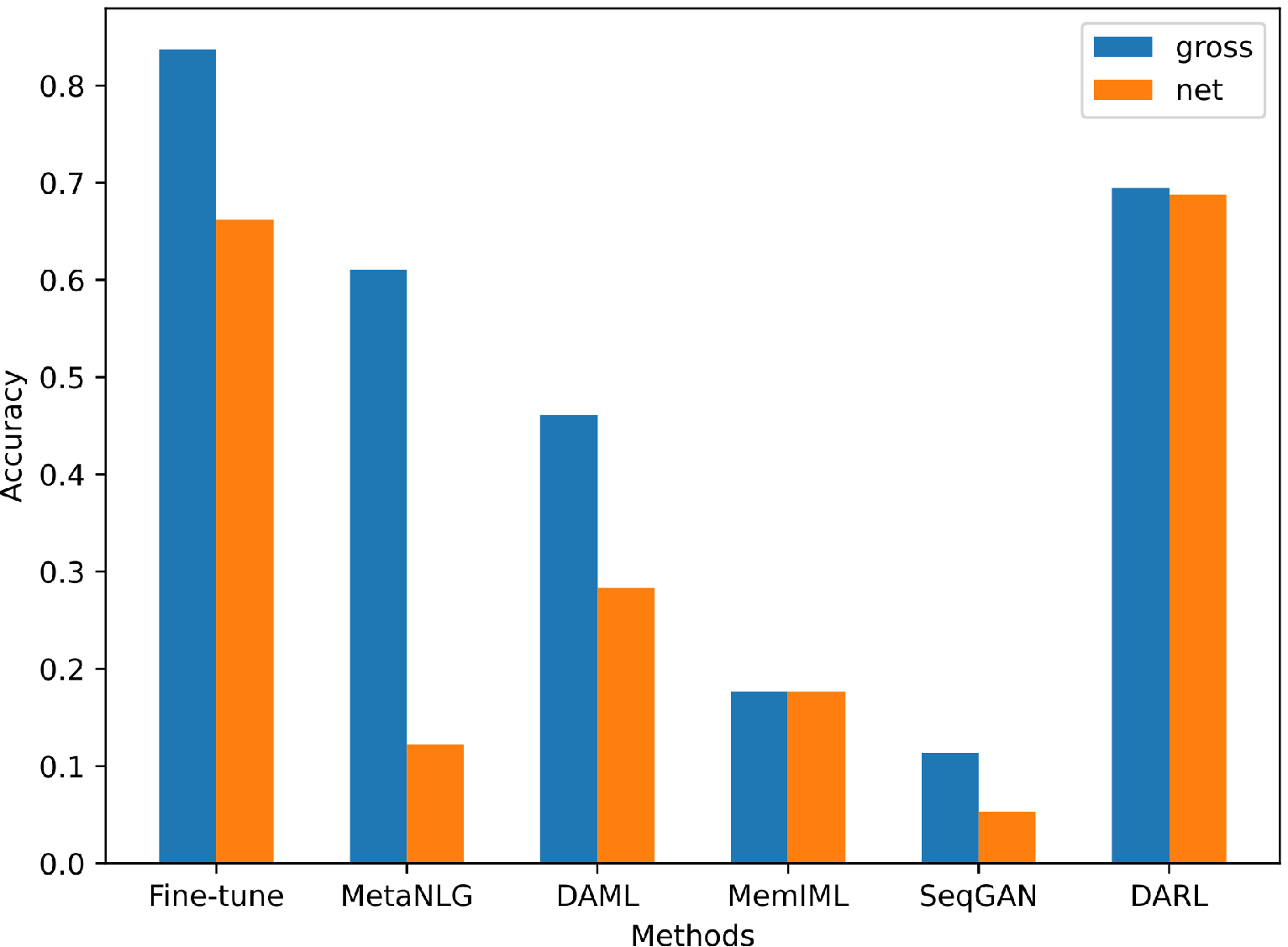}
	}
	\subfigure[office]{
		\includegraphics[width=0.3\textwidth]{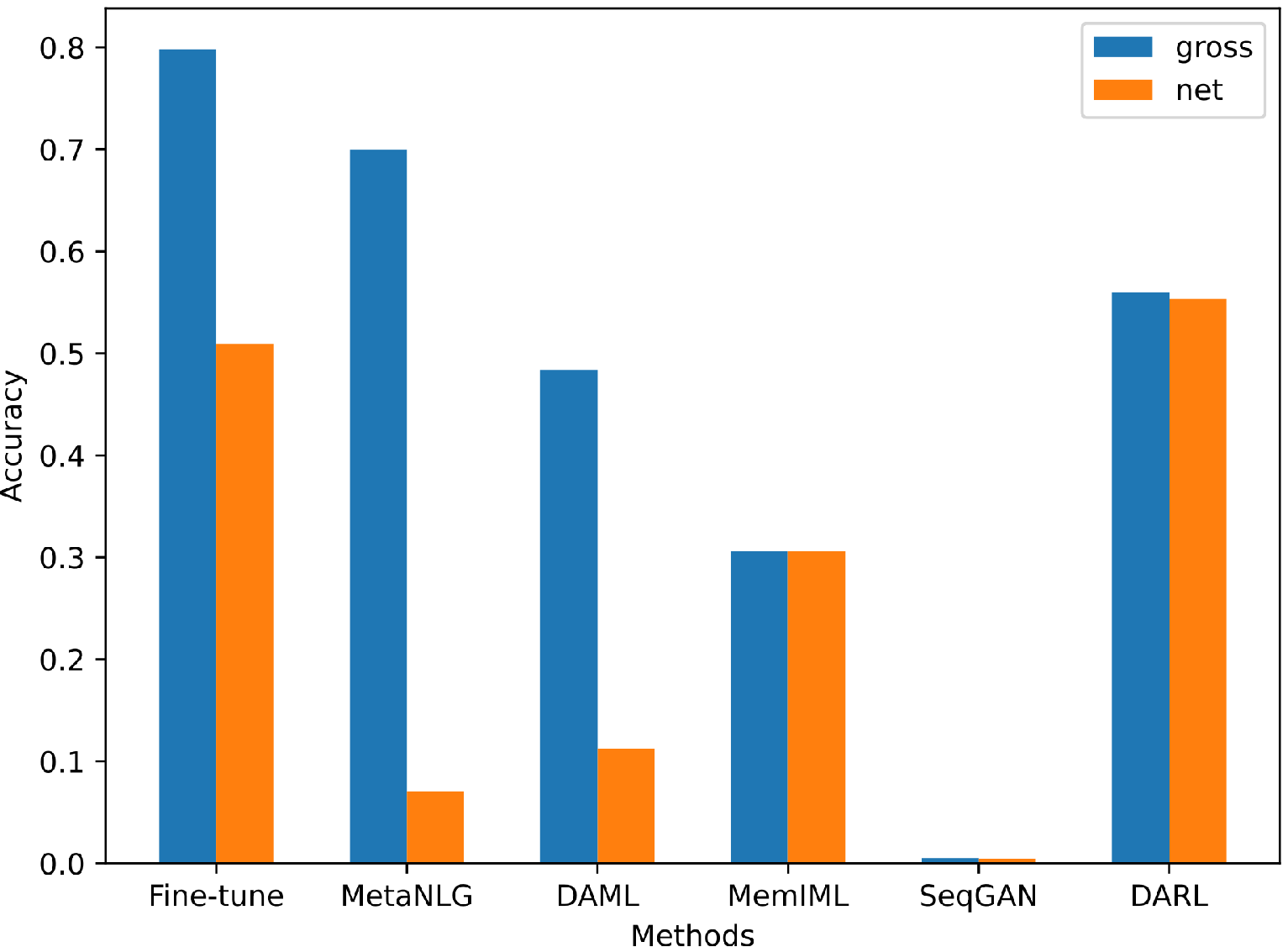}
	}
	\subfigure[phone]{
		\includegraphics[width=0.3\textwidth]{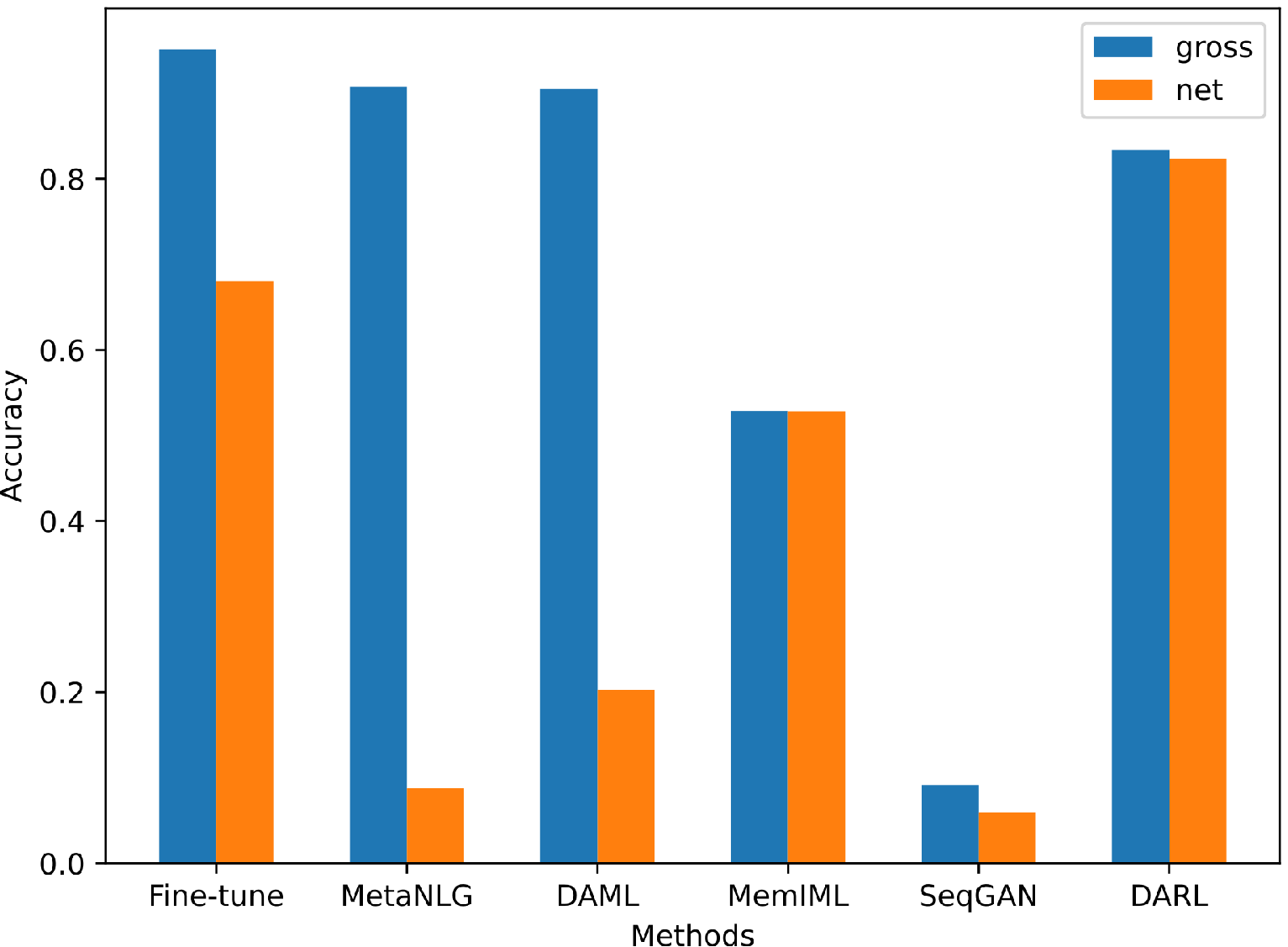}
	}
	\subfigure[tools]{
		\includegraphics[width=0.3\textwidth]{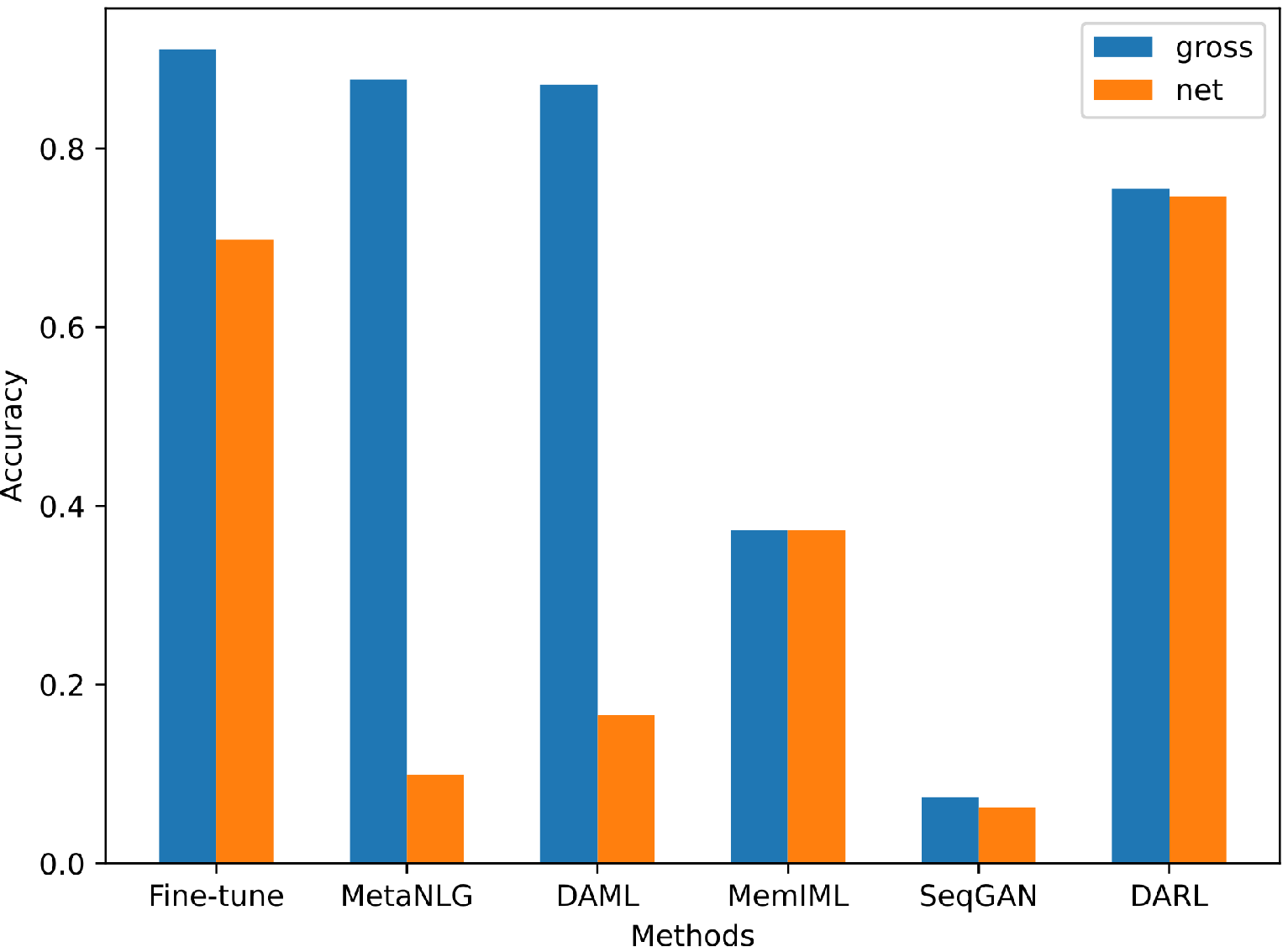}
	}
	\subfigure[average]{
		\includegraphics[width=0.3\textwidth]{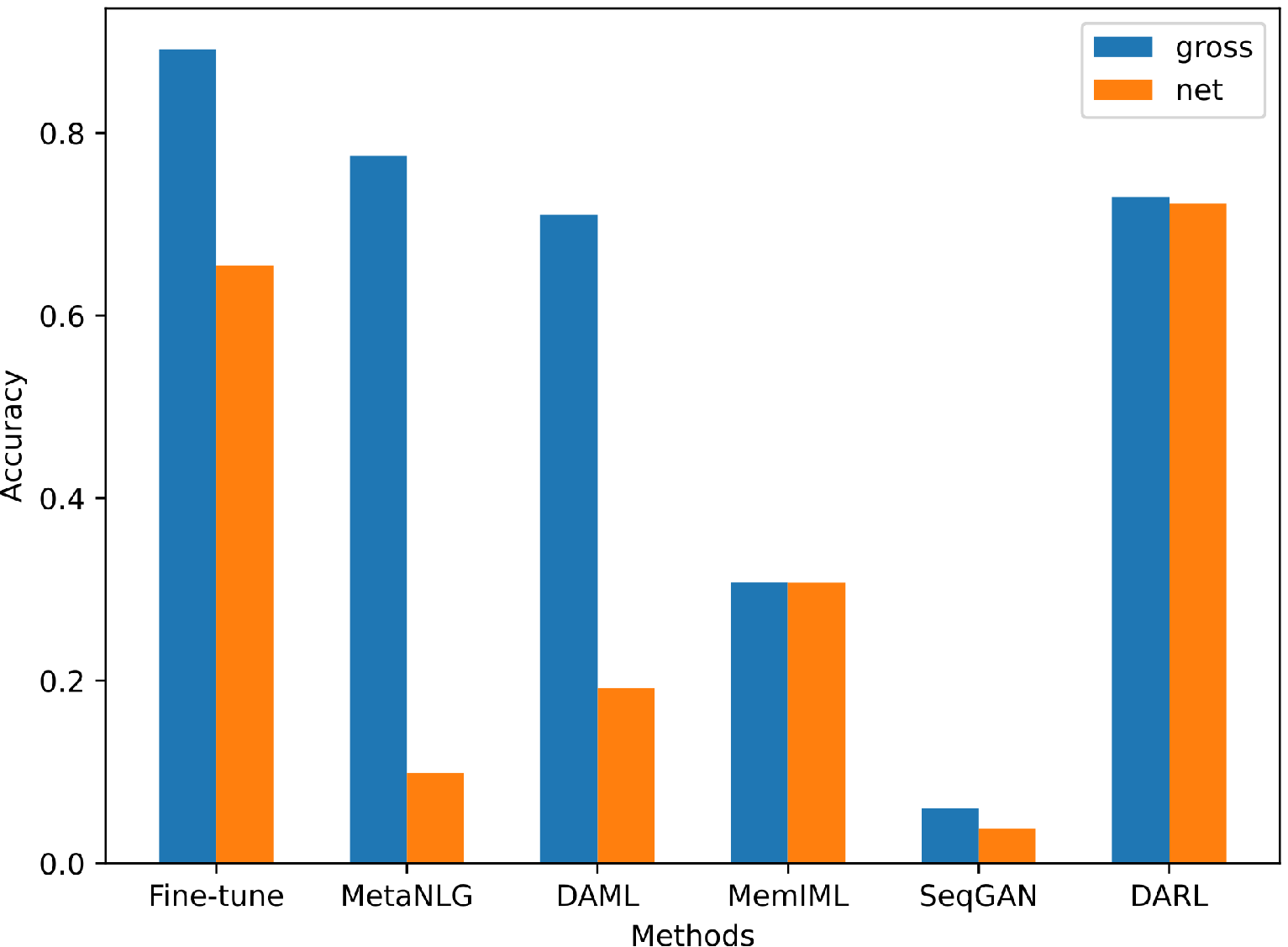}
	}
	\caption{The accuracy of generated texts on the 5-shot dataset. The mean value of the results across all target domains is shown in the subfigure titled average.}
	\label{fig.accuracy.5shot}
\end{figure}

\begin{figure}[H]
	\centering
	\subfigure[automotive]{
		\includegraphics[width=0.3\textwidth]{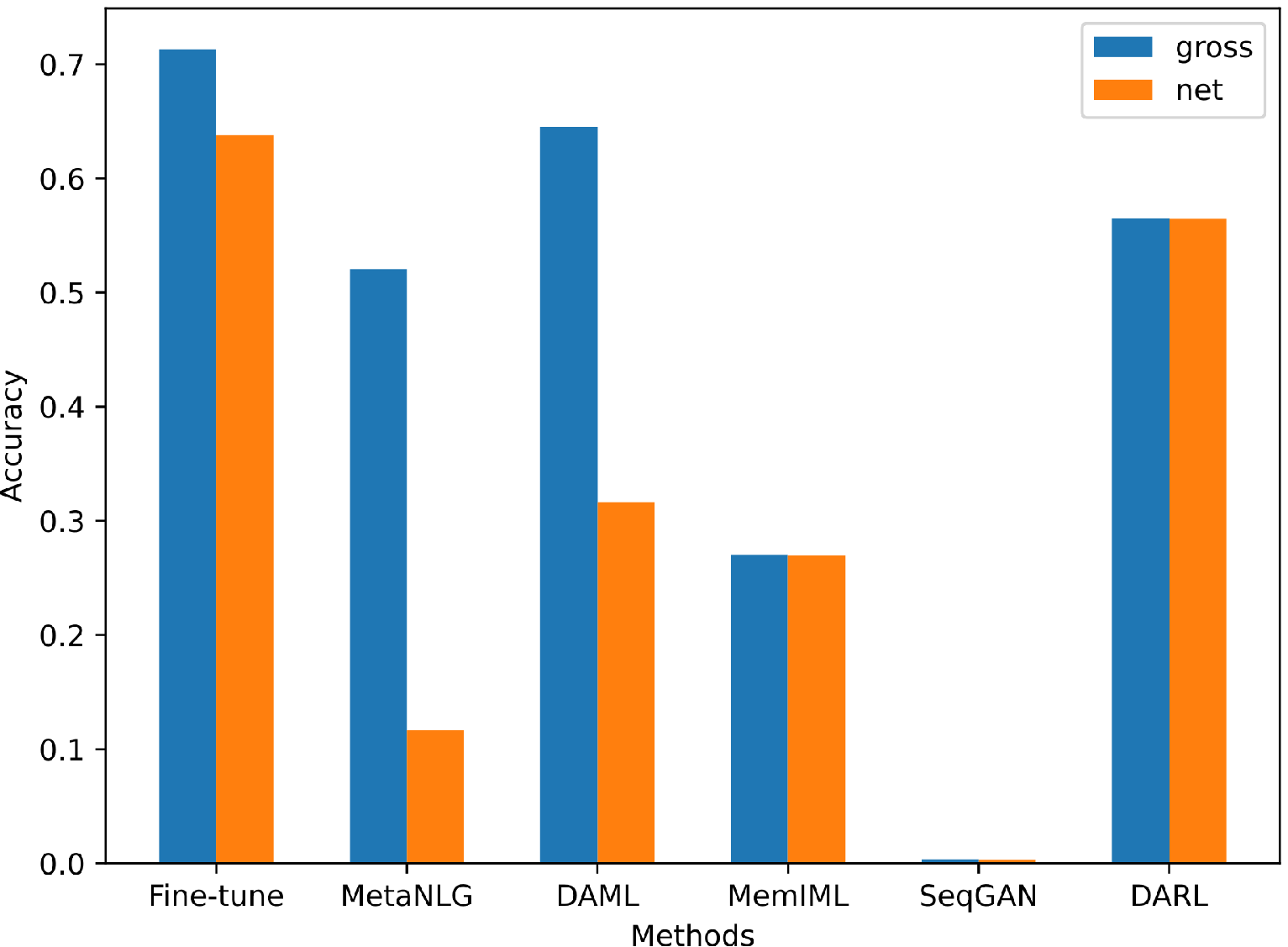}
	}
	\subfigure[music]{
		\includegraphics[width=0.3\textwidth]{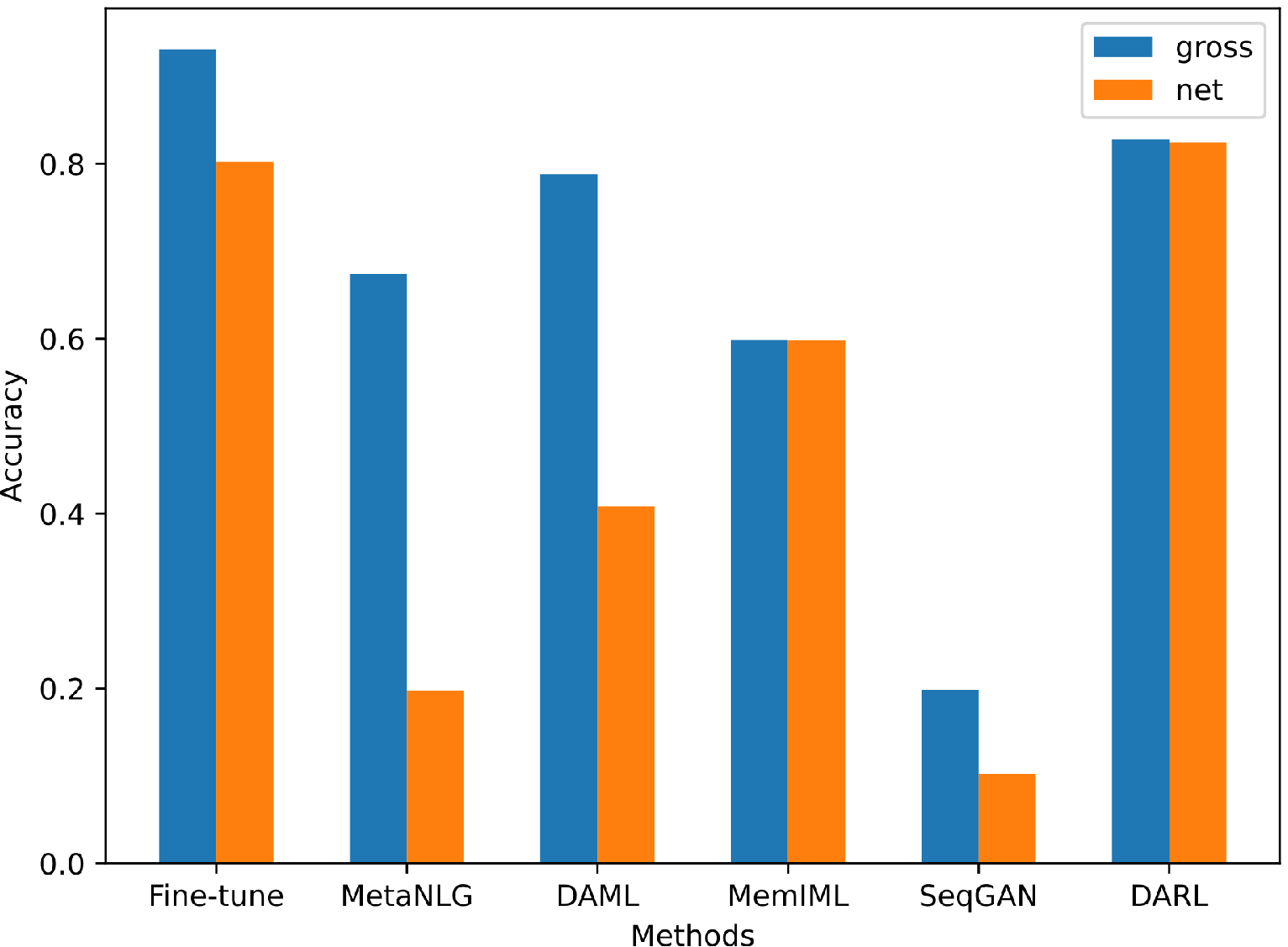}
	}
	\subfigure[office]{
		\includegraphics[width=0.3\textwidth]{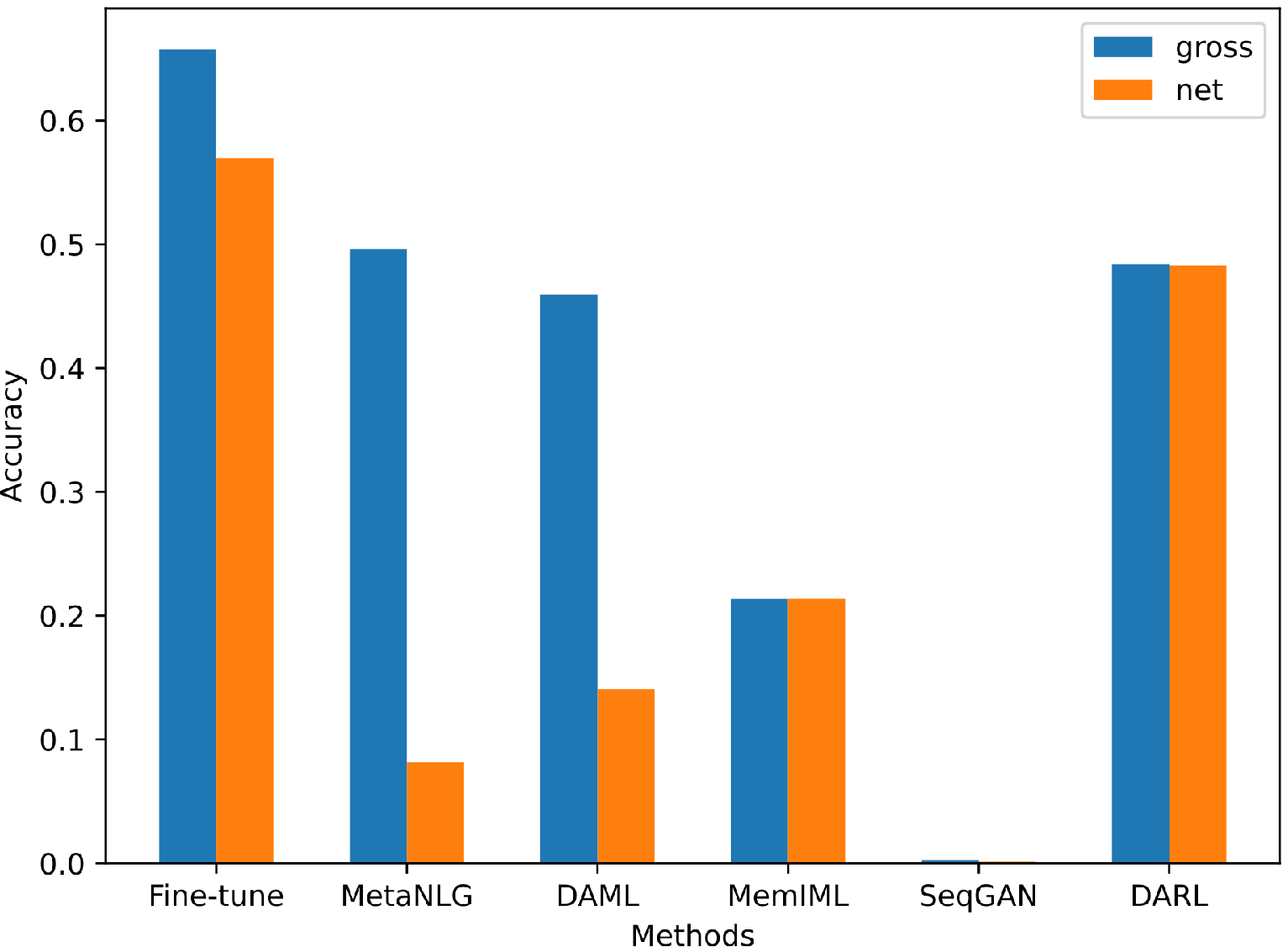}
	}
	\subfigure[phone]{
		\includegraphics[width=0.3\textwidth]{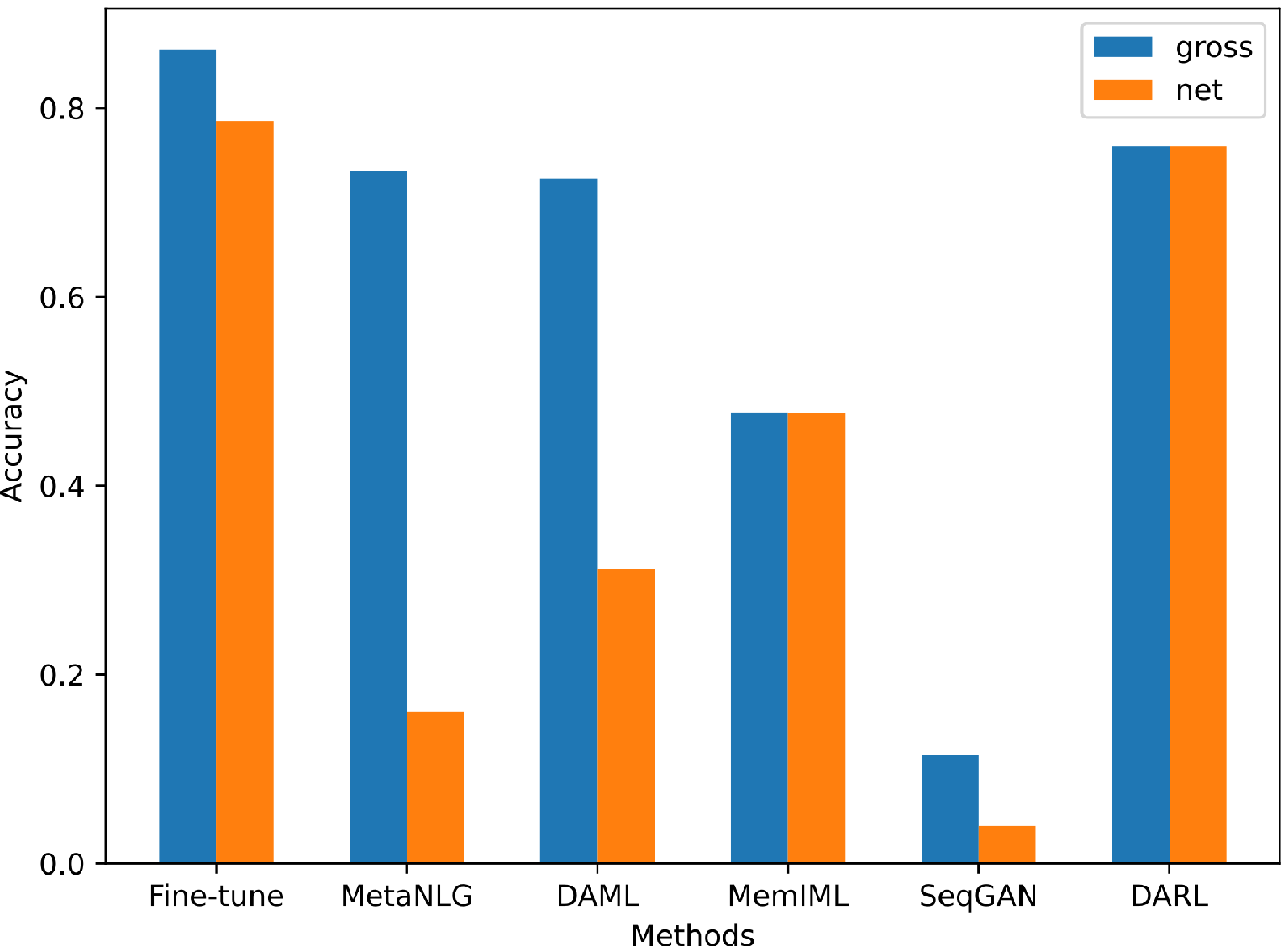}
	}
	\subfigure[tools]{
		\includegraphics[width=0.3\textwidth]{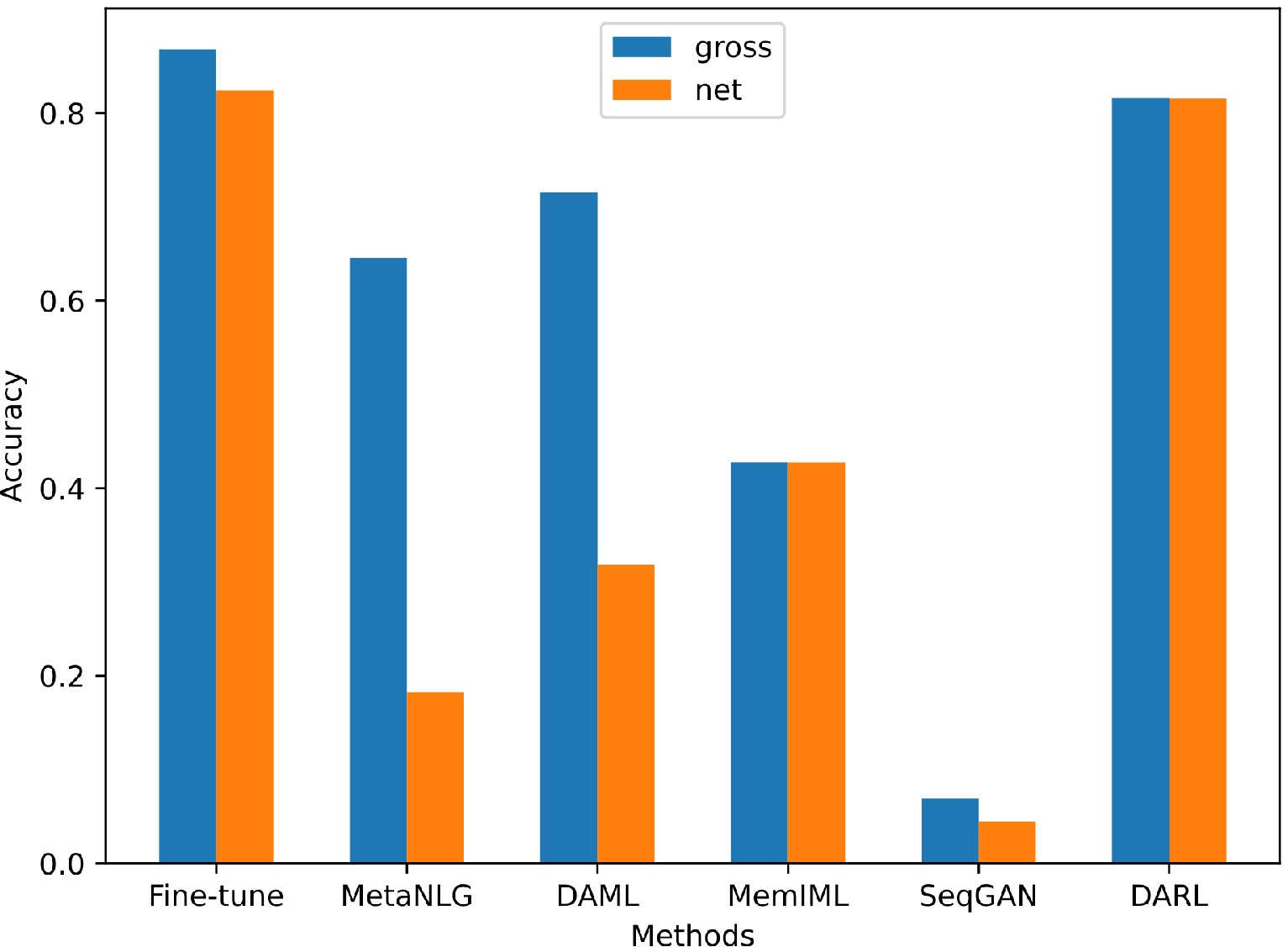}
	}
	\subfigure[average]{
		\includegraphics[width=0.3\textwidth]{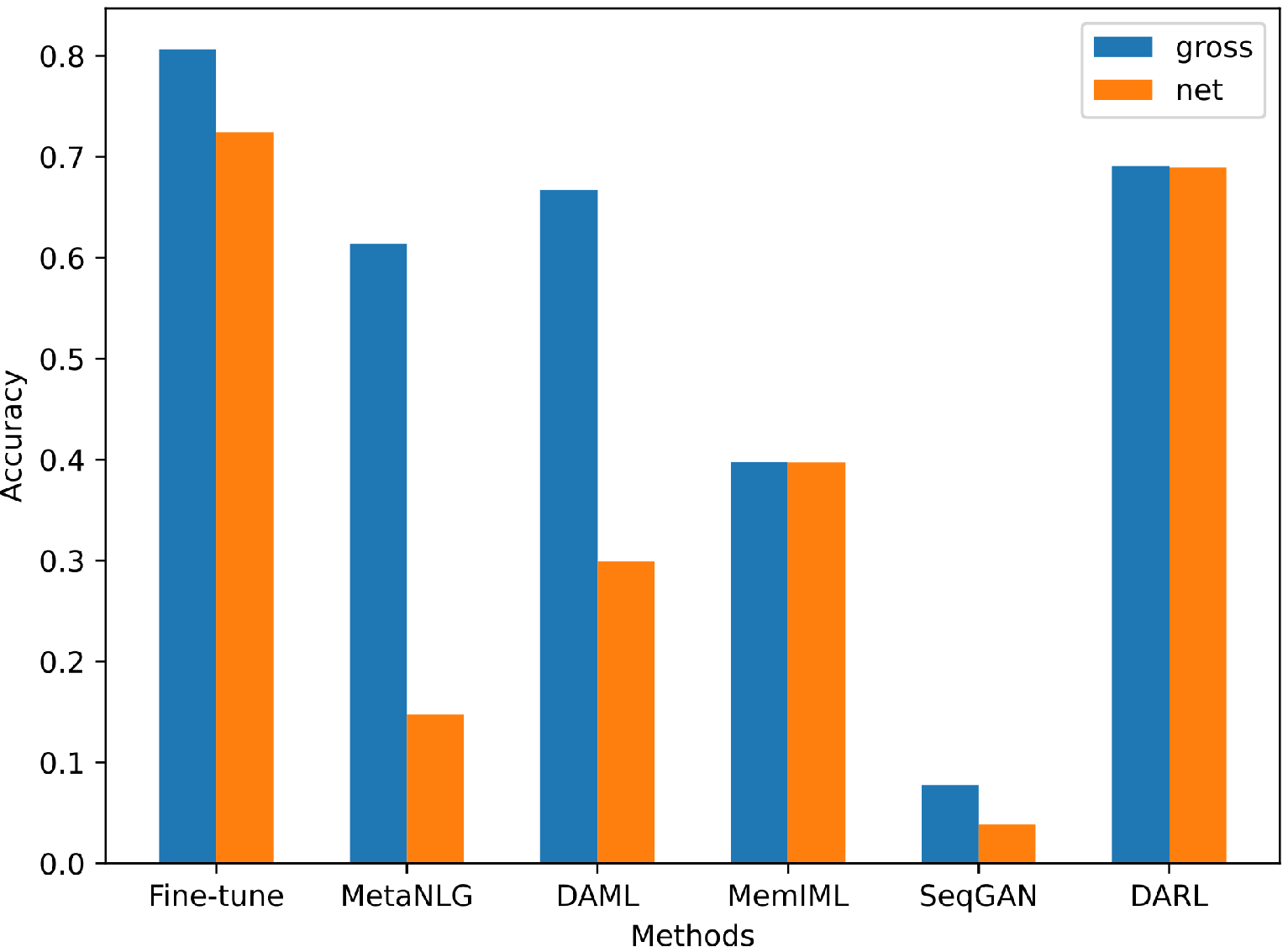}
	}
	\caption{The accuracy of generated texts on the 10-shot dataset. The mean value of the results across all target domains is shown in the subfigure titled average.}
	\label{fig.accuracy.10shot}
\end{figure}

The results are shown in Figure \ref{fig.accuracy.5shot} and Figure \ref{fig.accuracy.10shot}. In terms of gross accuracy, Fine-tune, MetaNLG, and DAML all have very notable advantages. But in terms of net accuracy, DARL and Fine-tune have significant advantages. Additionally, the data displays how different methods' variations in gross accuracy and net accuracy. The top three methods (Fine-tune, MetaNLG, and DAML) with the highest gross accuracy also suffered the greatest net accuracy declines. This implies that these 

\noindent methods generate more duplicate texts and consequently involve more tax penalties. The remaining three methods (DAML, MetaNLG, SeqGAN) only experience a slight decline in net accuracy, indicating that fewer duplicate texts are generated and that the effective output rate is higher. The phenomenon of high similarity between generated texts is consistent with the one mentioned in section \ref{sec_quality}. And it highlights shortcomings of the meta-learning-based method. Although overfitting can be reduced, this could also result in a lack of learning about the target task. For instance, MemIML performs better than the other two meta-learning-based methods in terms of overfitting, but it falls far short of DARL and Fine-tune in terms of representing the target domain.

In summary, the effective domain relevance of texts generated by DARL is higher than baselines. Additionally, DARL is no issue with overfitting.

\subsection{Analysis of hyperparameter $R$}

\begin{figure}[H]
	\centering
	\subfigure[average]{
		\includegraphics[width=0.3\textwidth]{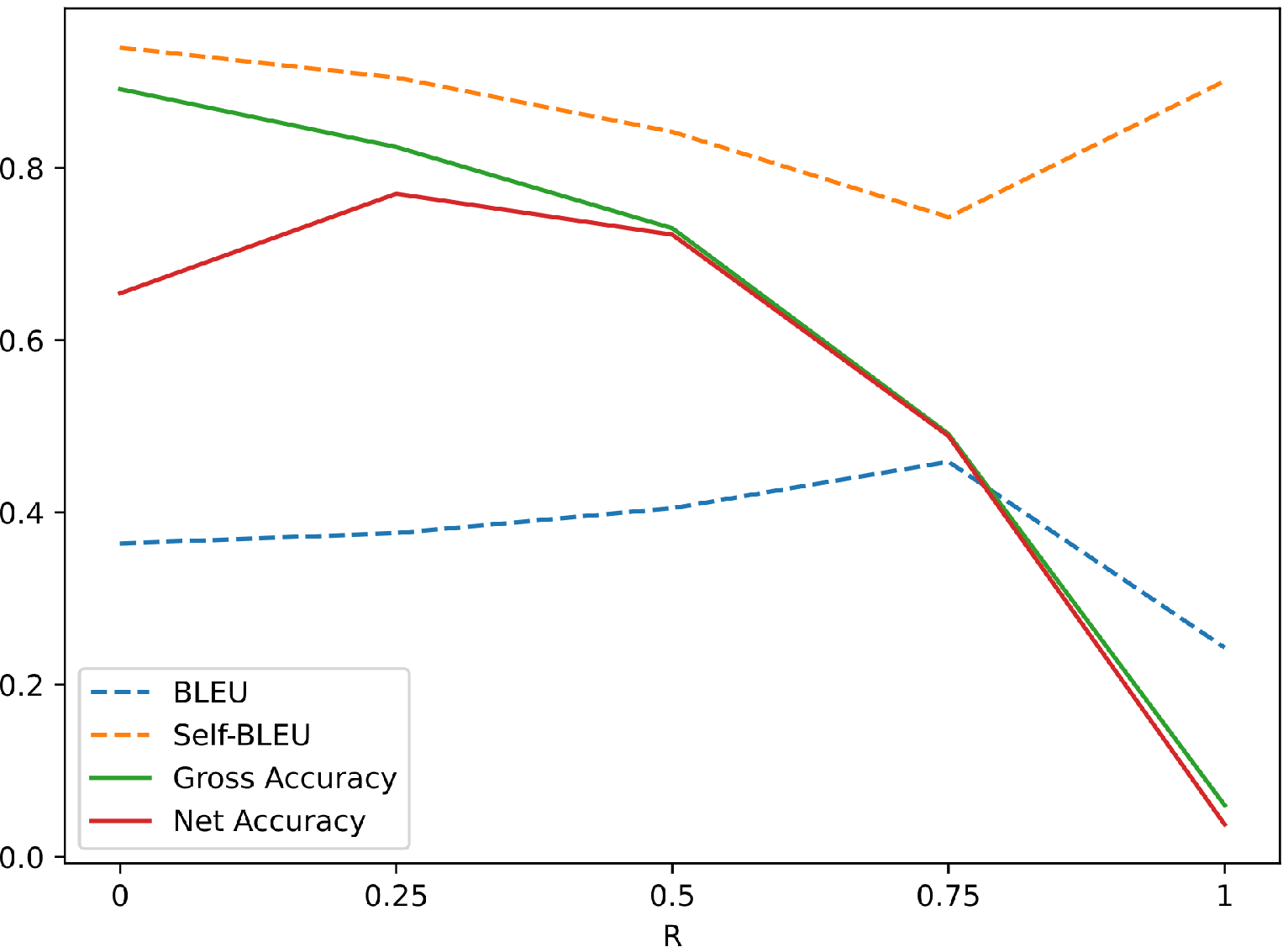}
	}
	\subfigure[automotive]{
		\includegraphics[width=0.3\textwidth]{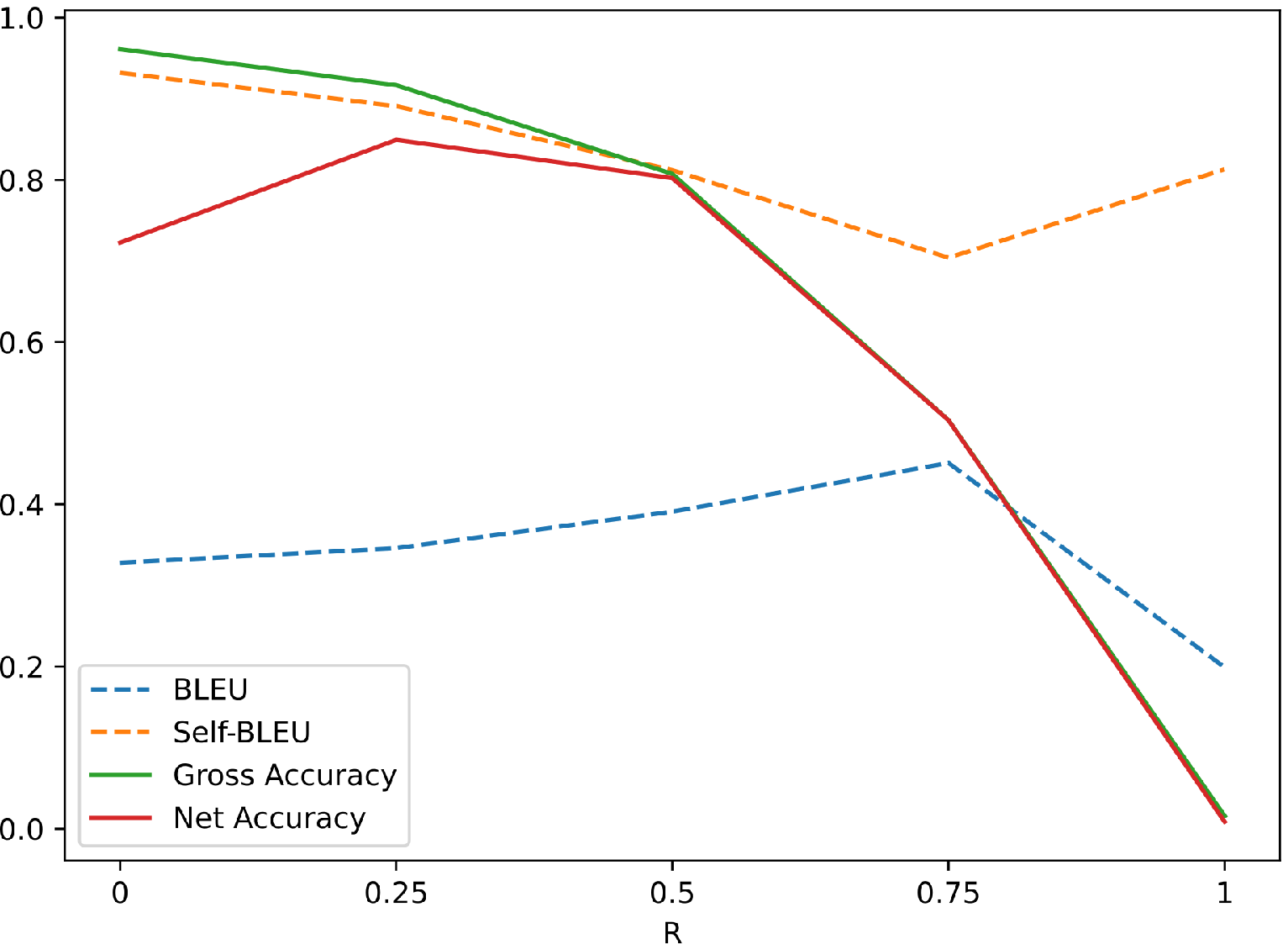}
	}
	\subfigure[music]{
		\includegraphics[width=0.3\textwidth]{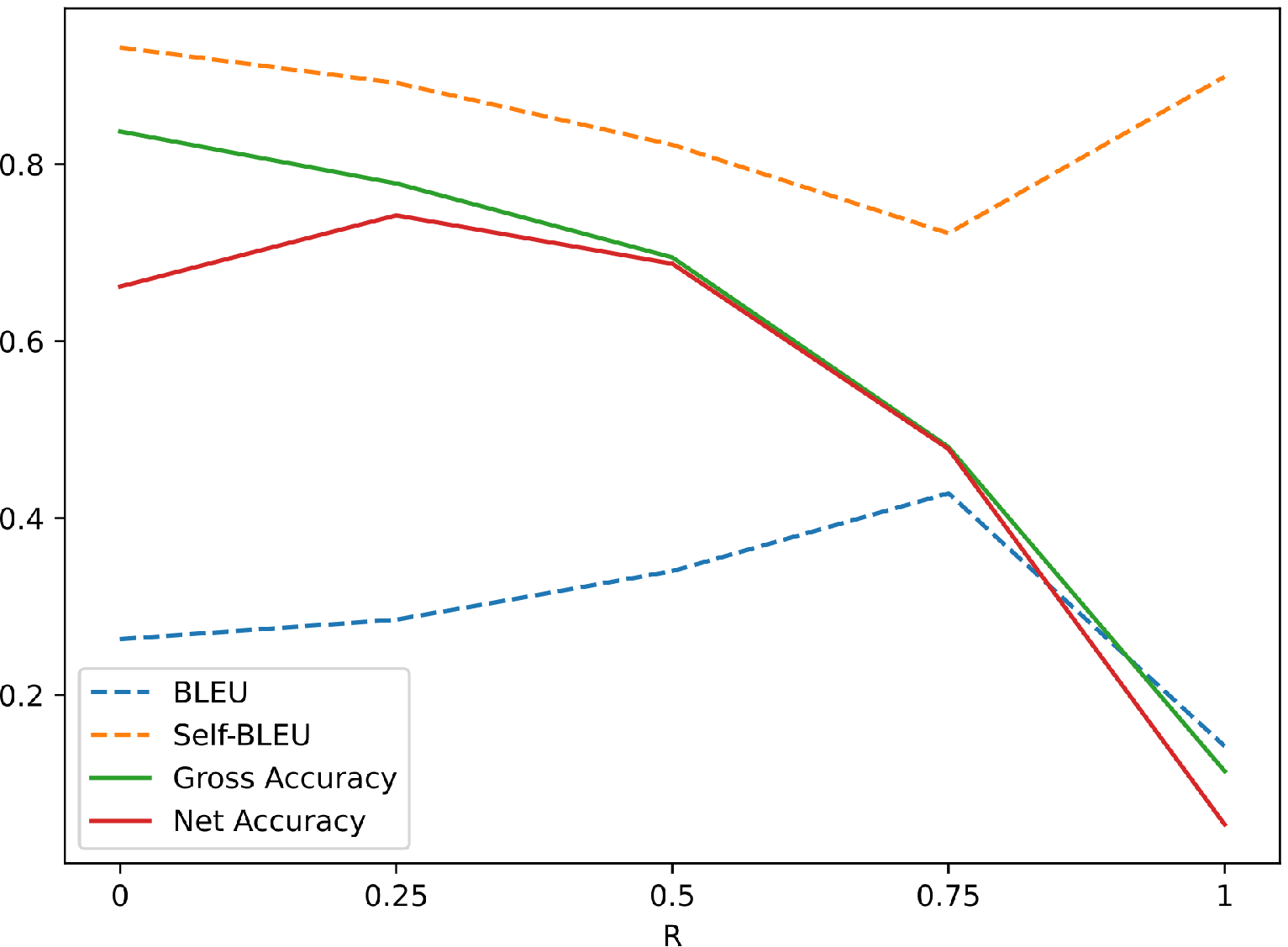}
	}
	\subfigure[office]{
		\includegraphics[width=0.3\textwidth]{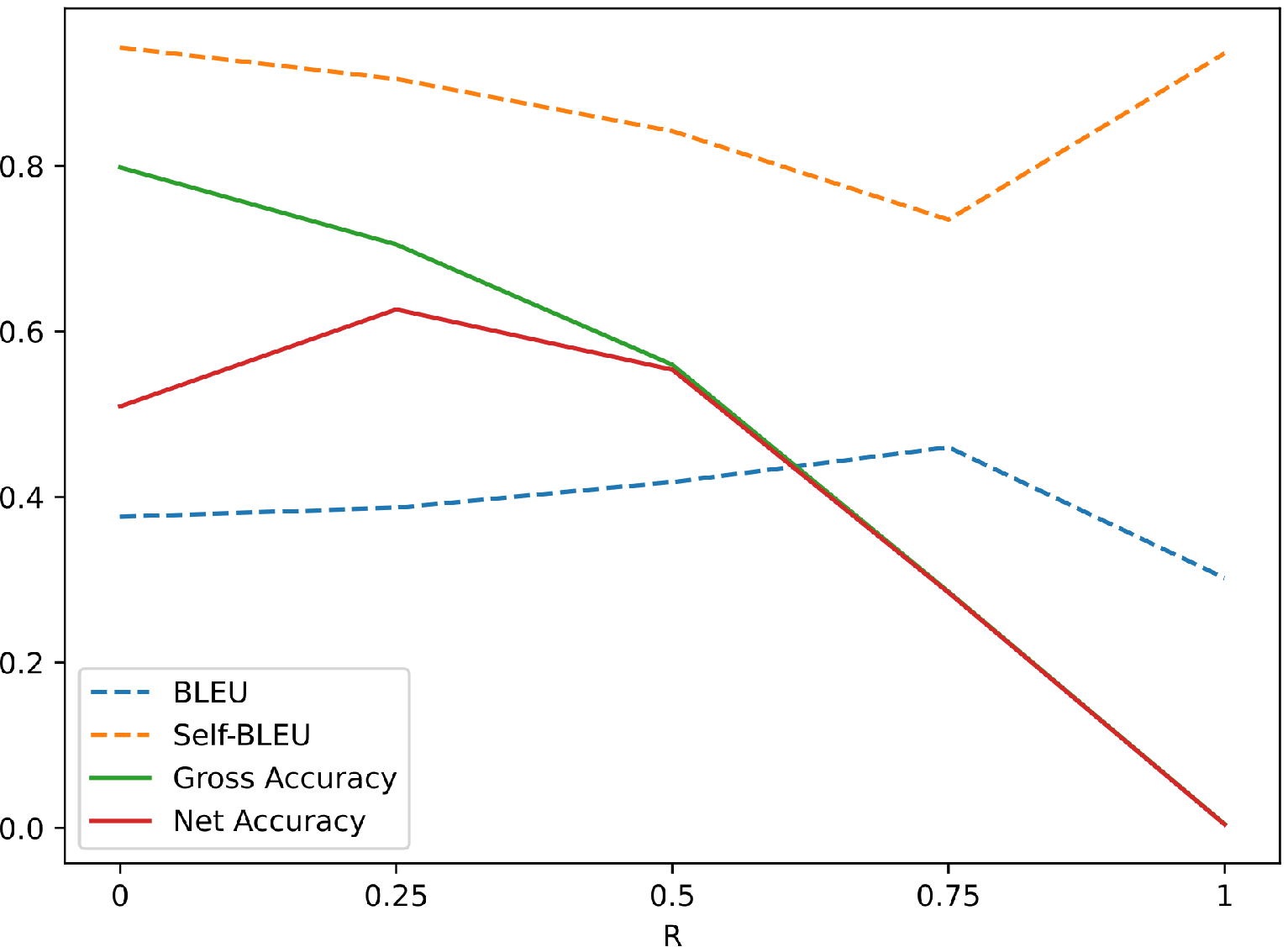}
	}
	\subfigure[phone]{
		\includegraphics[width=0.3\textwidth]{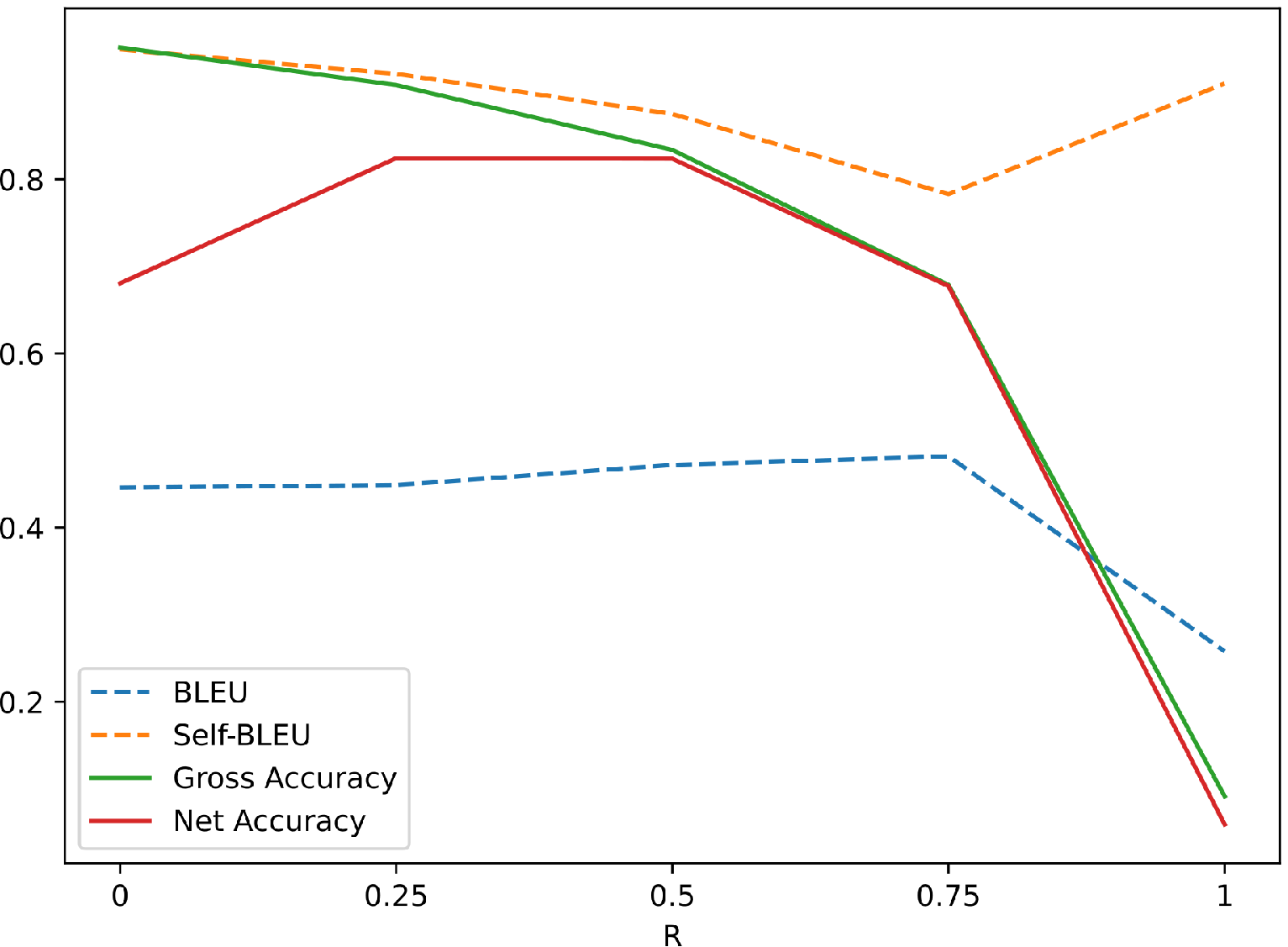}
	}
	\subfigure[tools]{
		\includegraphics[width=0.3\textwidth]{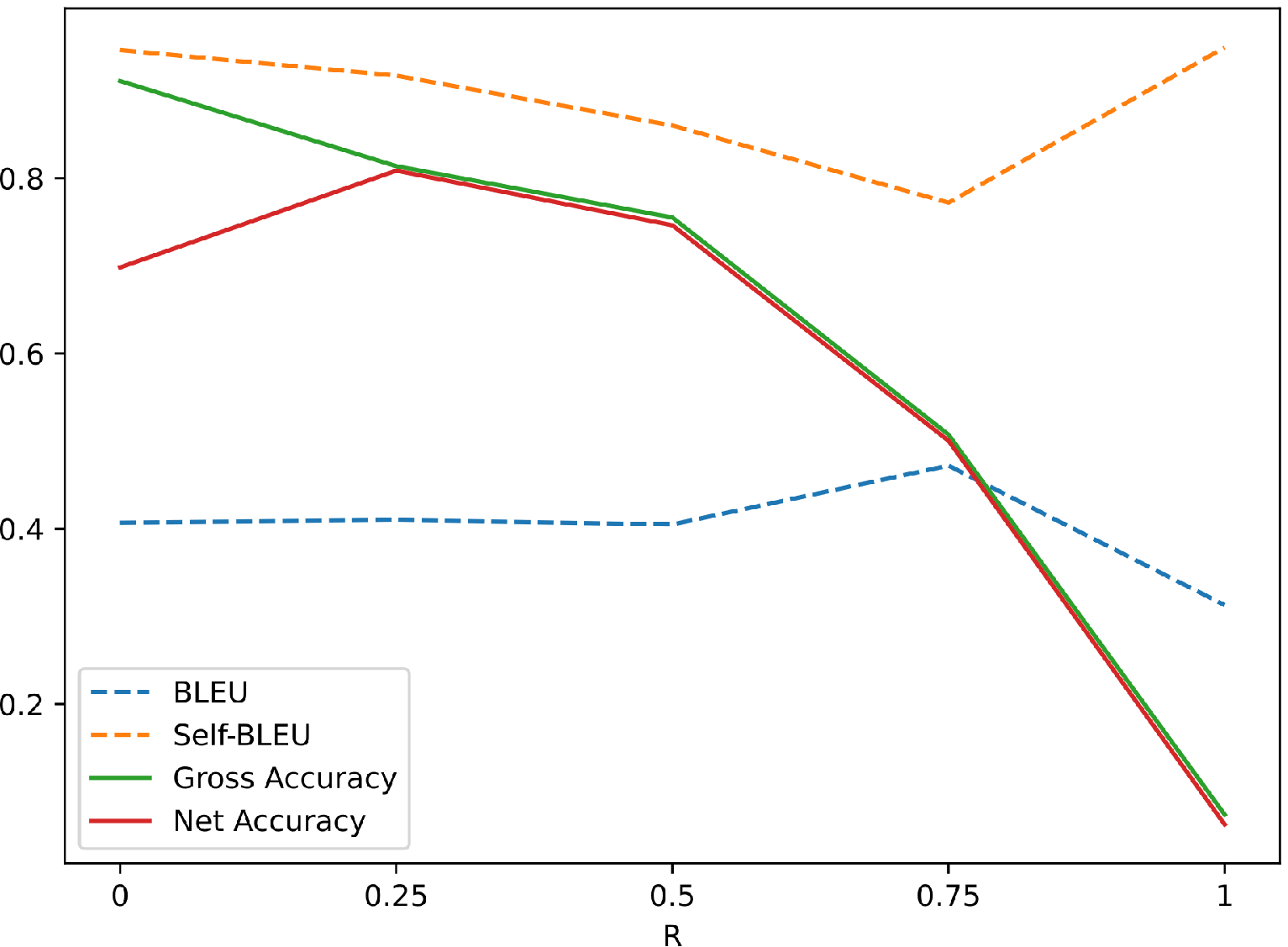}
	}
	\caption{The impact of $R$ on findings on the 5-shot dataset. The mean value of the results across all target domains is shown in the subfigure titled average.}
	\label{fig.r_5}
\end{figure}

\begin{figure}[h]
	\centering
	\subfigure[average]{
		\includegraphics[width=0.3\textwidth]{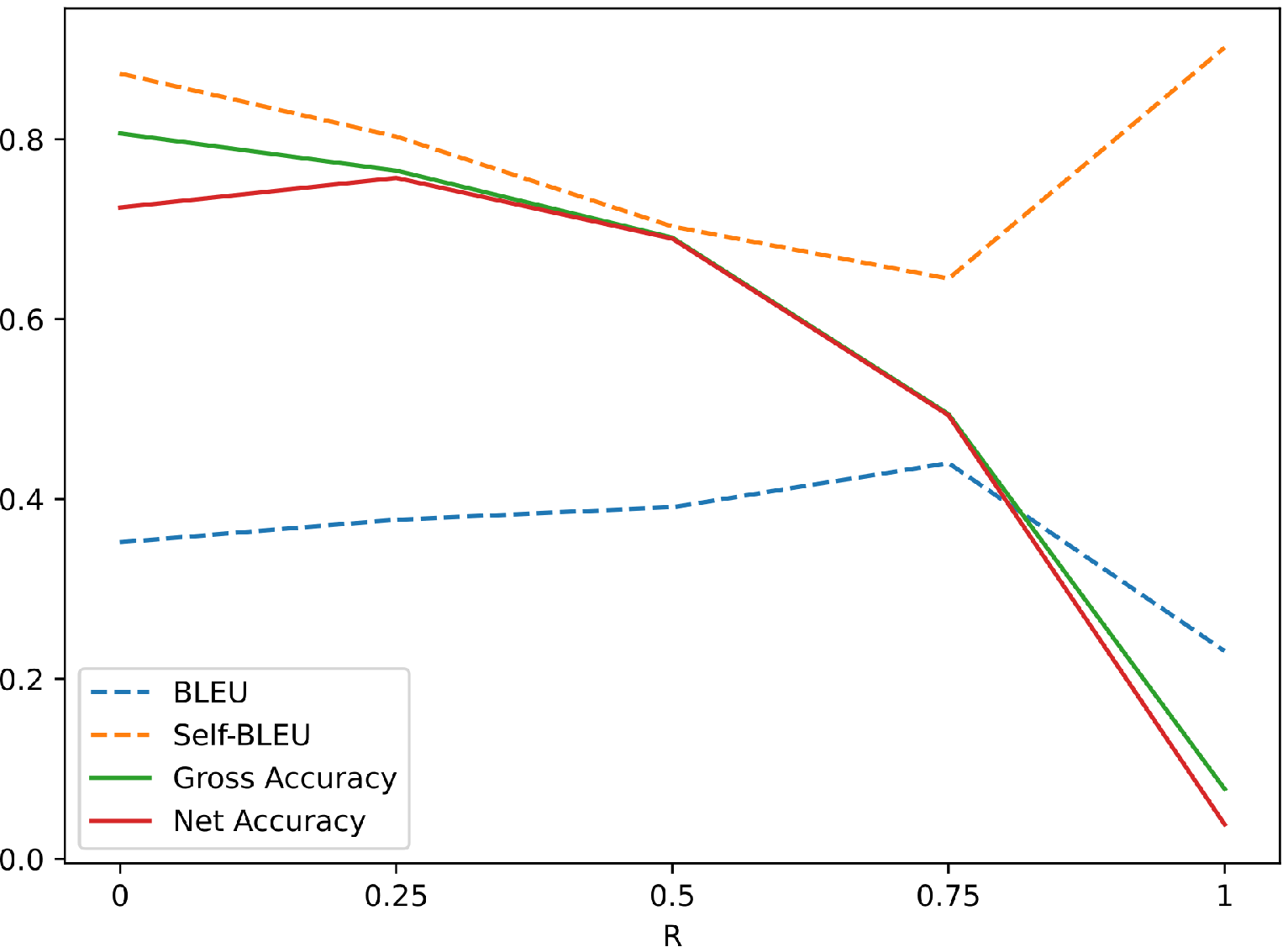}
	}
	\subfigure[automotive]{
		\includegraphics[width=0.3\textwidth]{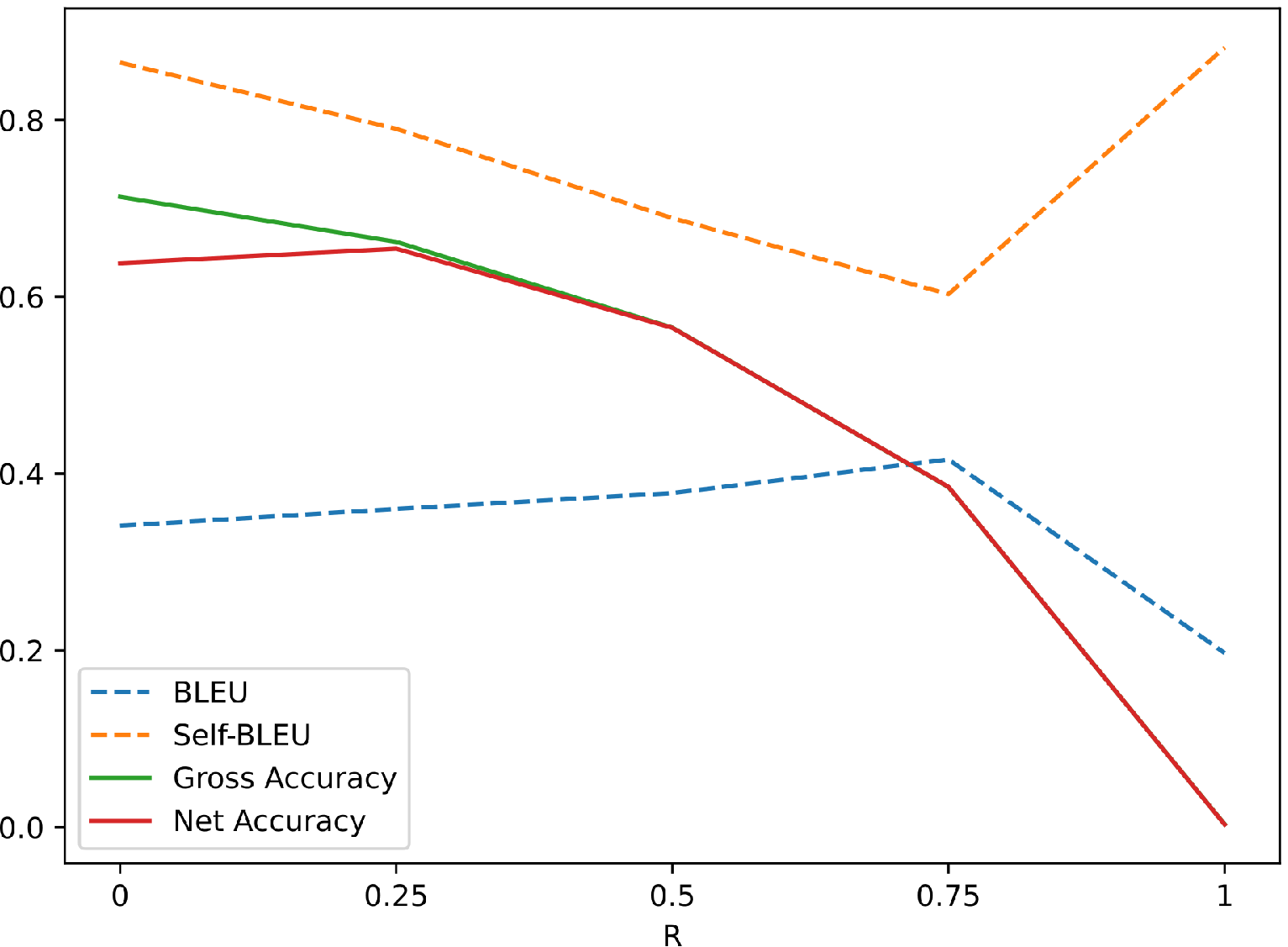}
	}
	\subfigure[music]{
		\includegraphics[width=0.3\textwidth]{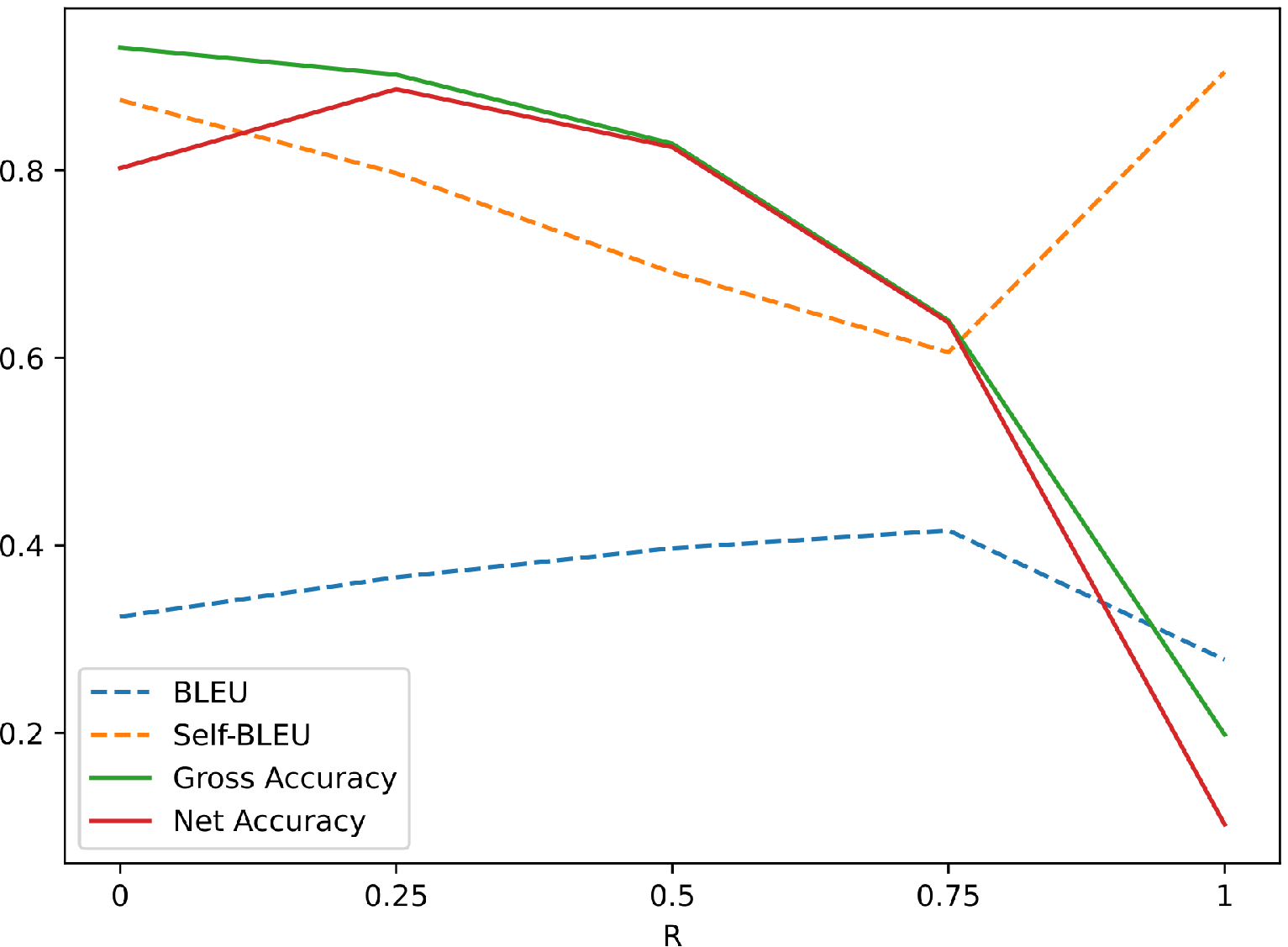}
	}
	\subfigure[office]{
		\includegraphics[width=0.3\textwidth]{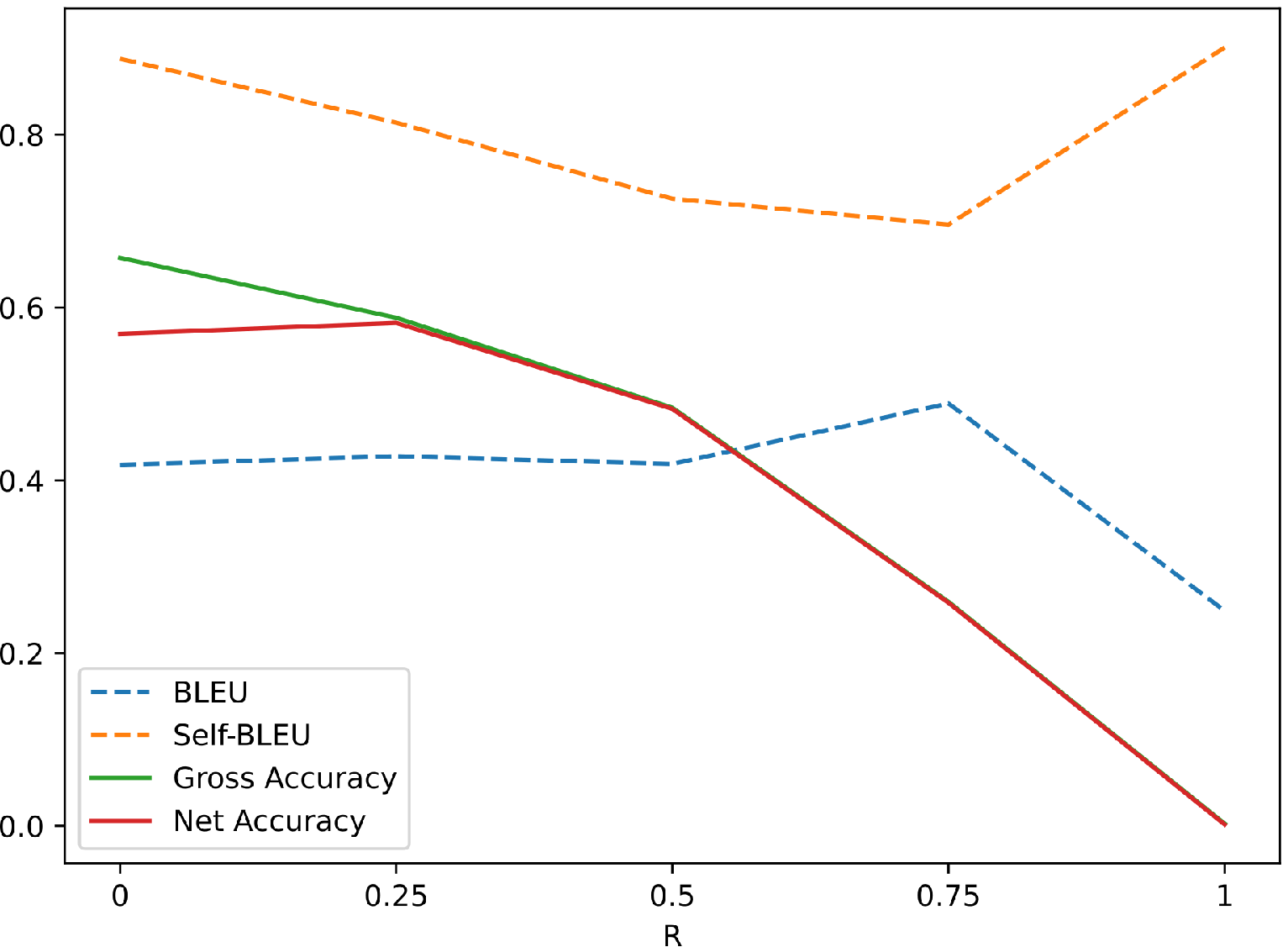}
	}
	\subfigure[phone]{
		\includegraphics[width=0.3\textwidth]{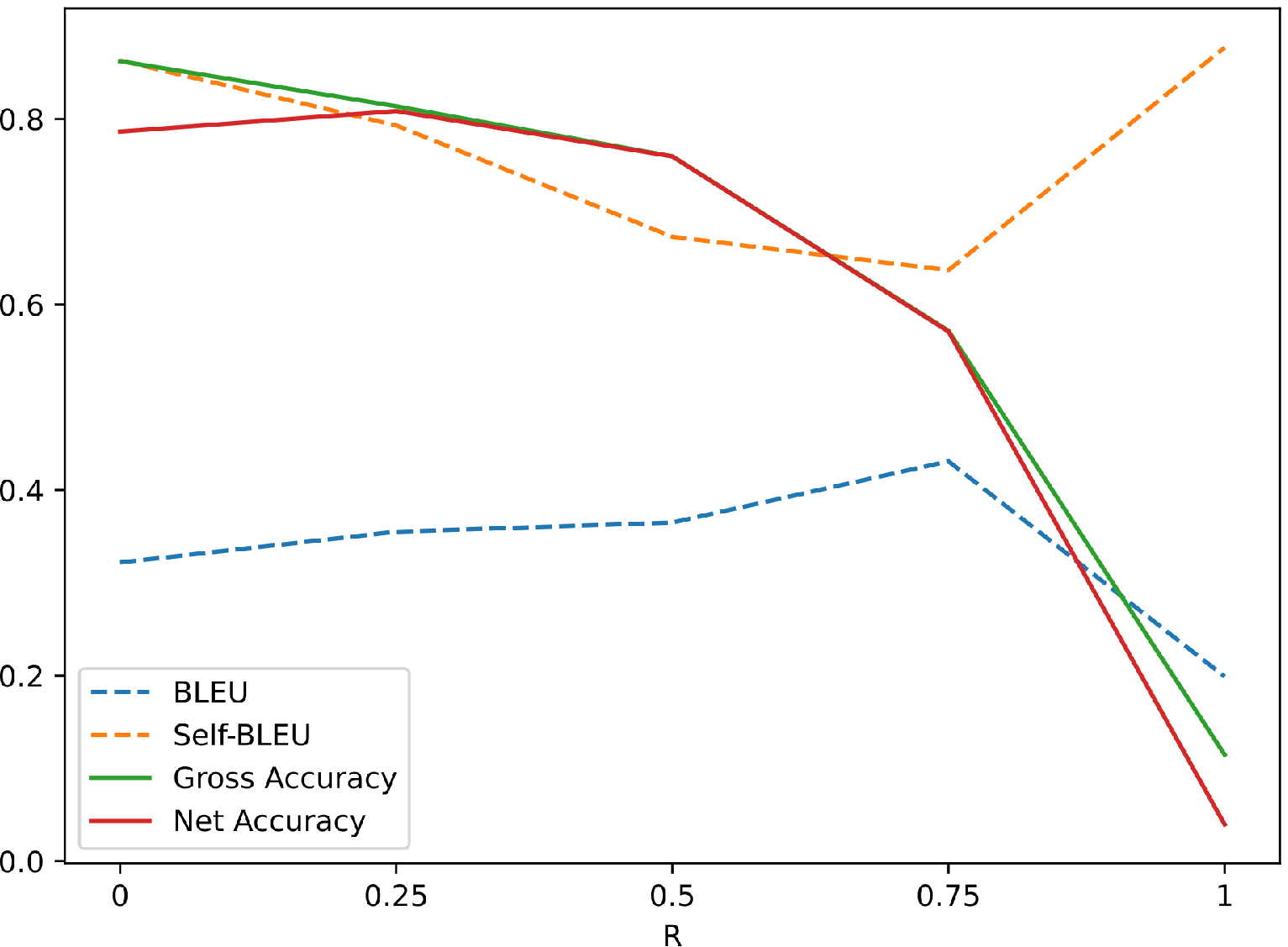}
	}
	\subfigure[tools]{
		\includegraphics[width=0.3\textwidth]{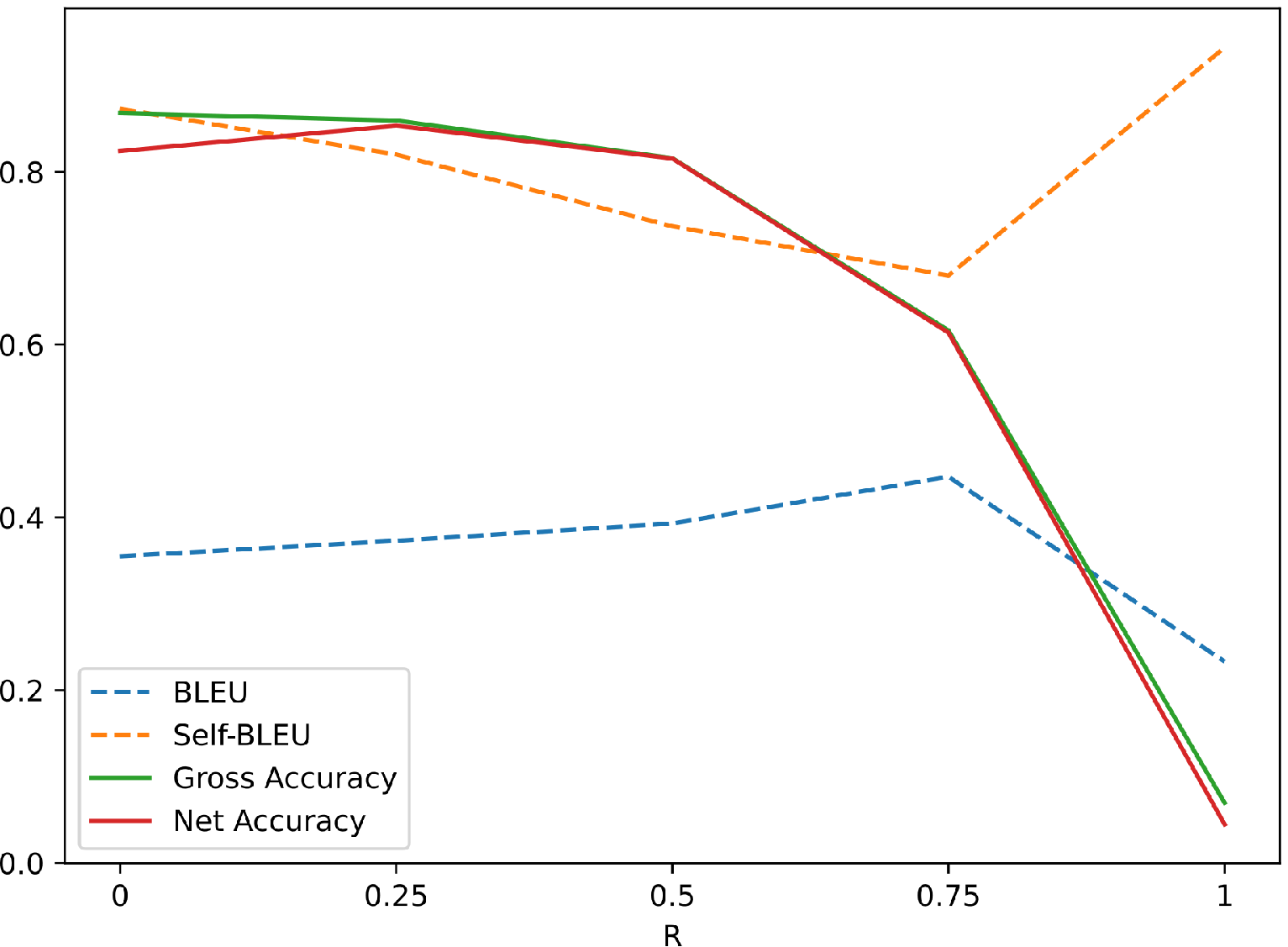}
	}
	\caption{The impact of $R$ on findings on the 10-shot dataset. The mean value of the results across all target domains is shown in the subfigure titled average.}
	\label{fig.r_10}
\end{figure}

To test the effect of $R$ on results, $R$ was set to be 0, 0.25, 0.5, 0.75, and 1, respectively. During the training process, as $R$ increases, fewer MLE is executed while more RL is executed. The results are shown in Figure \ref{fig.r_5} and Figure \ref{fig.r_10}. 

In terms of text quality, the fluency and diversity of the generated texts are two metrics that can be improved with more RL training, but maintaining some MLE can stop these two metrics from getting worse. The fluency and diversity metrics are at their worst when R is set to 1, no MLE is run, and DARL experiences severe pattern collapse. This once more demonstrates that the pattern collapse issue can be improved by including MLE in RL.

In terms of text domain relevance, gross accuracy decreases as RL training progresses, while net accuracy trends downward after a brief uptrend. When R is set to be 1, the fact that the accuracy of DARL is not zero supports the viability of changing the current probabilities to describe the target domain probabilities using RL. However, RL based on trial-and-error exploration is ineffective. Guided by reward signals rather than samples, the model is only capable of generating a very small amount of target domain texts. The addition of MLE significantly reduces this issue. With the introduction of MLE, DARL is now able to directly learn the target domain knowledge from samples. The difference between gross and net accuracy shows that DARL generates more duplicate texts when R is set to be a low value, proving that more MLE is not necessarily better.

As a result, the value of $R$ shouldn't be excessively high or low. When $R$ is low, although the generated texts' domain relevance is high, the quality is poor, and there are a lot of duplicate texts; When R is high, although the generated texts include fewer duplicate texts, the domain relevance is low, and there is a chance of pattern collapse, which could lead to poor text quality.

\section{Discussion}

In adversarial training, the discriminator is essential. The text classifier is used as a discriminator in text generation tasks based on adversarial training. It is the same as DARL. How will classifiers with different training methods affect the results? The comparison experiments were conducted to clarify this issue. In the experiments, the same CNN \cite{Kim2014} was trained by following three methods.

\begin{itemize}
	\item[$\bullet$] 
	\textbf{Classic}: All samples of a target domain (size is 2000) were positive samples and texts generated by DARL with same size were negative samples. The processes of pre-training and training were followed the work \cite{Yu2017} with $d-step=1$.   	
	\item[$\bullet$] 
	\textbf{Transition}: $k$ samples of a target domain were positive samples and texts generated by DARL with $k$ size were negative samples. The processes of pre-training and training were followed the classic method. And positive samples were kept same in each training.  
	\item[$\bullet$] 
	\textbf{Maml}: The processes of pre-training and training were followed the work of  MAML \cite{Finn2017}. In pre-training, CNN was trained with source domains. At the last iteration of pre-training, it was trained with the $k$ samples of target domain and the $k$ samples of DARL generated texts.
\end{itemize}

The benchmark used to compare the efficacy of training methods is the classic method that is successful and widely used in practice. As the results shown in Table \ref{tab.classifier_5} and Table \ref{tab.classifier_10}, the metrics of the classic method are measured values, and the metrics for the other two methods are the increase/decrease values compared to the classic method. It can be found that the measurements of the transition and maml methods do not differ significantly from those of the classic method, respectively. It is anticipated that using a small sample set to train the classifier using the classic method over several epochs would result in overfitting \cite{Hospedales2020}. However, in this task, half of the generated data in the sample set is changing. The classifier is motivated to continuously learn new information and new features as a result of the dataset's frequent updates, which prevents overfitting. This explains why the final results are not significantly affected by the selection of different training methods to update the discriminator parameters in the adversarial process.

In conclusion, when the sample size is small, it makes sense to use a classifier based on few-shot learning as a discriminator.

\begin{table}[H]
	\centering
	\caption{Impact evaluation of discriminator training methods using the classic method as a benchmark on the 5-shot dataset}
	\label{tab.classifier_5}
	\begin{tabular}{cccccc}
		\toprule
		Domains & Methods & BLEU & \tabincell{c}{Self\\BLEU} & \tabincell{c}{Gross\\Accuracy} & \tabincell{c}{Net\\Accuracy}\\
		\midrule
		\multirow{3}*{Automotive} & Classic & 0.340 & 0.769 & 0.796 & 0.795\\
		& Transition & 0.004 & 0.017 & 0.014 & 0.013\\
		& Maml & 0.015 & 0.017 & 0.016 & 0.015\\
		\hline
		\multirow{3}*{Music} & Classic & 0.306 & 0.789 & 0.689 & 0.687\\
		& Transition & 0.023 & 0.024 & 0.016 & 0.013\\
		& Maml & 0.029 & 0.016 & -0.004 & -0.005\\
		\hline
		\multirow{3}*{Office} & Classic & 0.368 & 0.810 & 0.543 & 0.541\\
		& Transition & 0.029 & 0.019 & 0.006 & 0.006\\
		& Maml & -0.014 & 0.010 & 0.006 & 0.005\\
		\hline
		\multirow{3}*{Phone} & Classic & 0.440 & 0.859 & 0.846 & 0.841\\
		& Transition & -0.009 & -0.001 & 0.021 & 0.023\\
		& Maml & -0.012 & 0.003 & 0.012 & 0.009\\
		\hline
		\multirow{3}*{Tools} & Classic & 0.395 & 0.840 & 0.748 & 0.743\\
		& Transition & 0.007 & 0.016 & 0.014 & 0.014\\
		& Maml & 0.022 & 0.016 & -0.004 & -0.003\\
		\hline
		\multirow{3}*{Average} & Classic & 0.370 & 0.814 & 0.724 & 0.721\\
		& Transition & 0.011 & 0.014 & 0.015 & 0.014\\
		& Maml & 0.003 & 0.017 & 0.005 & 0.005\\
		\bottomrule
	\end{tabular}
\end{table}

\begin{table}[H]
	\centering
	\caption{Impact evaluation of discriminator training methods using the classic method as a benchmark on the 10-shot dataset}
	\label{tab.classifier_10}
	\begin{tabular}{cccccc}
		\toprule
		Domains & Methods & BLEU & \tabincell{c}{Self\\BLEU} & \tabincell{c}{Gross\\Accuracy} & \tabincell{c}{Net\\Accuracy}\\
		\midrule
		\multirow{3}*{Automotive} & Classic & 0.351 & 0.652 & 0.573 & 0.573\\
		& Transition & 0.010 & 0.013 & -0.001 & -0.001\\
		& Maml & -0.010 & 0.004 & 0.009 & 0.009\\
		\hline
		\multirow{3}*{Music} & Classic & 0.356 & 0.645 & 0.838 & 0.836\\
		& Transition & -0.001 & 0.013 & 0.007 & 0.007\\
		& Maml & -0.026 & -0.002 & 0.014 & 0.014\\
		\hline
		\multirow{3}*{Office} & Classic & 0.391 & 0.684 & 0.477 & 0.477\\
		& Transition & -0.007 & 0.016 & 0.000 & -0.001\\
		& Maml & 0.007 & 0.023 & 0.027 & 0.027\\
		\hline
		\multirow{3}*{Phone} & Classic & 0.364 & 0.668 & 0.756 & 0.755\\
		& Transition & 0.021 & 0.021 & 0.007 & 0.007\\
		& Maml & -0.024 & -0.003 & 0.025 & 0.025\\
		\hline
		\multirow{3}*{Tools} & Classic & 0.352 & 0.702 & 0.811 & 0.811\\
		& Transition & 0.013 & 0.018 & 0.016 & 0.015\\
		& Maml & 0.000 & 0.008 & 0.010 & 0.010\\
		\hline
		\multirow{3}*{Average} & Classic & 0.363 & 0.670 & 0.691 & 0.690\\
		& Transition & 0.007 & 0.016 & 0.006 & 0.006\\
		& Maml & -0.011 & 0.006 & 0.017 & 0.017\\
		\bottomrule
	\end{tabular}
\end{table}

\section{Conclusion}
To generate domain texts with few samples, an RL-based framework called DARL is proposed in this paper. The fundamental concept behind this study is to use RL to change the distribution of words that can convey the target domain. Additionally, MLE training is incorporated into RL to improve sample utilization, which also alleviates the pattern collapse problem. Due to the limited size of the dataset, a classifier based on few-shot learning is used as the discriminator. Extensive experiments demonstrate the efficacy of DARL. 

However, it is discovered in some experiments that DARL tends to learn local knowledge of particular samples. The explanation of this phenomenon helps to improve the learning ability of DARL by constructing special samples. In future work, this question will continue to be investigated by related studies.

\section*{Acknowledgments}	
This work was supported by the Key Research and Development Program of Sichuan Province (Grant no.2021YFG0156).

\bibliographystyle{unsrt}
\bibliography{darl}

\end{document}